%% file: main.tex
\documentclass[runningheads]{llncs}

\usepackage{acronym}
\usepackage{amssymb}
\usepackage{tikz}
\usepackage{pgfplots}
\usepackage{pgfplotstable}
\usepackage{verbatim}
\usepackage{subcaption}
\usepackage{amsmath}
\usepackage{pifont}
\usepackage{subcaption}
\usepackage[hidelinks, hyperfootnotes=false]{hyperref} 
\usepackage{color}

\usepackage{dirtytalk}
\usepackage{booktabs}
\usepackage{multirow}
\usepackage{makecell}
\usepackage{graphicx}
\usepackage{svg}
\usepackage{nicematrix}
\usepackage{makecell}
\usepackage{amsmath}
\usepackage{pifont}
\usepackage{orcidlink}   

\usepackage{xcolor}
\usepackage{xspace}
\usepackage{tcolorbox} 
\usepackage{xcolor, soul}
\usepackage{colortbl}
\definecolor{mygreen}{rgb}{0,0.5,0} 
\definecolor{reallylightgray}{rgb}{0.95,0.95,0.95} 
\newcommand{\todo}[1][]{\textcolor{red}{\textbf{X}}}
\usetikzlibrary{positioning, shapes.geometric, arrows.meta, fit, calc}
\pgfplotsset{compat=1.18}
\usepgfplotslibrary{fillbetween}   
\usetikzlibrary{intersections}
\usepackage{pgfplots}
\pgfplotsset{compat=1.18} 
\usepackage[T1]{fontenc}

\usepackage{graphicx,verbatim}

\usepackage{xfrac}
\usepackage{diagbox}
\newcommand{\counts}[2]{{\footnotesize\sfrac{#1}{#2}}}

\begin{document}

\acrodef{CT}{Computed Tomography}
\acrodef{MCC}{Matthews Correlation Coefficient}
\acrodef{MoE}{Mixture-of-Experts}
\acrodef{MR}{Magnetic Resonance}
\acrodef{US}{Ultrasound}

\definecolor{TUMBlue}{HTML}{0065BD}
\definecolor{TUMSecondaryBlue}{HTML}{005293}
\definecolor{TUMSecondaryBlue2}{HTML}{003359}
\definecolor{TUMBlack}{HTML}{000000}
\definecolor{TUMWhite}{HTML}{FFFFFF}
\definecolor{TUMDarkGray}{HTML}{333333}
\definecolor{TUMGray}{HTML}{808080}
\definecolor{TUMLightGray}{HTML}{CCCCC6}
\definecolor{TUMAccentGray}{HTML}{DAD7CB}
\definecolor{TUMAccentOrange}{HTML}{E37222}
\definecolor{TUMAccentGreen}{HTML}{A2AD00}
\definecolor{TUMAccentLightBlue}{HTML}{98C6EA}
\definecolor{TUMAccentBlue}{HTML}{64A0C8}

\definecolor{PlotDeepBlue}{HTML}{005CAA}
\definecolor{PlotSkyBlue}{HTML}{56B4E9}
\definecolor{PlotLightGreen}{HTML}{3F8F4C}
\definecolor{PlotTeal}{HTML}{1B9E9A}
\definecolor{PlotGoldenOrange}{HTML}{E69F00}
\definecolor{PlotVermillion}{HTML}{D55E00}
\definecolor{PlotPurple}{HTML}{7570B3}

\definecolor{GSMoEYellow}{HTML}{FFD86F} 
\definecolor{GSMoEOrange}{HTML}{FF8C42} 
\definecolor{GSMoRed}{HTML}{C92A2A} 
\definecolor{ModelUltramarine}{HTML}{6FD7E0} 
\definecolor{Horizon}{HTML}{1381BF} 
\definecolor{Cerulean}{HTML}{123272} 
\definecolor{Endeavour}{HTML}{0756A7}
\definecolor{TUMDarkGray}{HTML}{333333}

\newcommand{\cmark}{\ding{51}}%
\newcommand{\xmark}{\ding{55}}%
%

\title{Generalist-Specialist Mixture-of-Experts for Rare Pathology Detection in Multimodal Imaging}
\titlerunning{GS-MoE}
\author{Johannes Kaiser \inst{1, 2, *}\orcidlink{0009-0007-0819-8751} \and
Florian Braunmiller \inst{1, *} \and
Daniel Rückert \inst{1, 3}\orcidlink{0000-0002-5683-5889} \and
Georgios Kaissis \inst{2}\orcidlink{0000-0001-8382-8062} 
}

\institute{Chair for AI in Healthcare and Medicine, Technical University of Munich (TUM) and TUM University Hospital, Munich, Germany \and
Hasso Plattner Institute for Digital Engineering, University of Potsdam, Potsdam, Germany \and
Department of Computing, Imperial College London, UK
}

\authorrunning{J. Kaiser et al.}

  
\maketitle              

\begin{abstract}
AI models for multimodal medical imaging must balance modality-specific specialization with cross-modal shared representations, a trade-off that pure \ac{MoE} architectures currently fail to satisfy.
Expert-based routing improves in-domain learning but may sacrifice cross-modal signals, which appear particularly important for rare (low-prevalence) pathologies in our experiments.
To resolve this, we introduce \textit{Generalist-Specialist-MoE} (\textit{GS-MoE}), a two-branch \ac{MoE} architecture that couples a cross-modal generalist model with distinct modality-specific specialists (experts) via domain-constrained feature fusion.
On RadImageNet (1.35M images, 165 pathologies, three modalities), GS-MoE recovers detection of six low-prevalence pathologies on which every baseline scores F1 $=$ 0, with per-class gains up to +0.60 F1.
It attains this while even slightly exceeding dense and specialist-only MoE aggregate baselines (MCC 0.770), while using ${\sim}53\%$ fewer active parameters at inference than the strongest 
investigated dense model.
\end{abstract}


\begingroup
\renewcommand{\thefootnote}{}
\footnotetext{Code is available at
  \href{https://github.com/Johannes-Kaiser/Generalist_Specialist_MoE}
  {\textcolor{blue}{\underline{GitHub}}}.\\ 
  * Equal contribution authors}
\endgroup

\section{Introduction}

Medical imaging analysis has shifted from small, single-organ cohorts to large, multimodal, multi-organ datasets such as RadImageNet~\cite{mei2022radimagenet}, comprising 1.35 million instances across 165 pathologies of varying prevalence.
While such large-scale datasets enable learning generalized representations, monolithic architectures struggle to reconcile the heterogeneity
~\cite{lakkapragada2022mitigatingnegativetransfermultitask}, arising naturally from variations in patient demographics, anatomy/pathology, and imaging modalities.

Jointly training a single model across diverse imaging modalities exposes a fundamental conflict between two competing transfer effects.
\textit{Negative transfer}~\cite{jiang2023forkmerge,wang2019characterizing} occurs when conflicting gradients across weakly related modalities hinder learning, degrading performance on low-prevalence data groups \cite{neg_3_fu2025rdam}.
\textit{Positive transfer} arises when shared anatomical structures provide a synergistic signal, benefiting low-prevalence classes with limited in-domain data  \cite{pos_verma2023longtailchest,pos_huang2023_mtl}.

To benefit low-prevalence data groups (i.e., classes with few training samples we use as a data-driven proxy for \textit{rarity}, without implying clinical rarity), this motivates a clear desideratum: an architecture must simultaneously minimize negative transfer effects while retaining the potential benefits of cross-modal positive transfer.
Unfortunately, this cannot be resolved trivially, as avoiding negative transfer requires training on highly related data, while positive transfer requires incorporating unrelated data to augment the limited in-domain information on low-prevalence pathologies.

To address this conflict, we propose the use of \textit{Generalist-Specialist Mixture-of-Experts (GS-MoE)}, a dual-stream \ac{MoE} architecture that combines a modality-gated \ac{MoE} with experts acting as specialists on a related subset of the data with a global generalist trained on all available multimodal data. \acp{MoE} address this by distributing data across distinct experts, operating on well-aligned subsets, while still being capable of processing diverse, multimodal datasets.

\acp{MoE} have gained substantial traction driven by their ability to scale model capacity to trillions of parameters \cite{fedus2022switch,68_jiang2024mixtralexperts,lepikhin2020gshard} with minimal computational overhead by activating only a sparse subset of experts \cite{15_mu2026comprehensivesurveymixtureofexpertsalgorithms,16_fedus2022reviewsparseexpertmodels,39_Cai_2025,45_gan2025mixtureexpertsmoebig}. 
The modularity of \acp{MoE} enables GS-MoE to scale to large datasets by splitting data into well-aligned subsets with domain-specific experts. In vision, prior work focuses mostly on \ac{MoE} vision transformers with patch-wise experts \cite{39_Cai_2025,51_han2024vimoeempiricalstudydesigning,55_riquelme2021scalingvisionsparsemixture,77_videau2025mixtureexpertsimageclassification,78_10377734}, but also extends to CNN-based instance-wise routing \cite{wang2019} and interpretability \cite{55_riquelme2021scalingvisionsparsemixture}.

We build on instance-wise routing \cite{wang2019,65_chopra2025medmoemodalityspecializedmixtureexperts} by routing inputs based on the imaging modality.
Like uni-modal monolithic architectures, modality-gated \acp{MoE} cannot benefit from positive transfer learning.
To address this shortcoming and fulfill our aforementioned desideratum, unlike prior works in \ac{MoE}-based medical image classification \cite{123_kanwal2024equippingcomputationalpathologysystems,105_li2024m4multiproxymultigatemixture}, segmentation \cite{109_ou2022patcher,110_chen2024low,111_wang2024sam}, and VLMs \cite{114_jiang2024med,64_jiang2024m4oe}, we introduce an additional \textit{generalist branch} alongside the \ac{MoE}. Conceptually, the \ac{MoE} provides modality-specific latent encodings of the input, which are augmented by shared cross-modal infor\allowbreak mation from a general branch to benefit from multimodal transfer learning \cite{omnismola}. 
Our contributions are as follows: 
\begin{enumerate}
    \item We identify cross-domain gradient interference as a likely driver of low-prevalence collapse under modality routing, and propose GS-MoE, a sparse dual-stream architecture that isolates it while retaining positive transfer.
    \item We make six low-prevalence pathologies detectable that every dense and specialist-only MoE baseline fails entirely (F1 $=$ 0) with gains up to $+0.60$ F1, attributed to cross-modal positive transfer via the generalist.
    \item These tail gains come without aggregate cost, as GS-MoE also leads all baselines in overall MCC (0.770) while using ${\sim}53\%$ fewer active parameters at inference than the strongest dense baseline, at matched total parameters.
\end{enumerate}
We expect the modular design of GS-MoE to enable scalability across new tasks, modalities, and patient demographics. 

\section{Methods}\label{sec:methods}
\subsection{Routing Strategies: Implicit vs.\ Explicit}
\label{sec:methods_routing_strategies}
multimodal classification induces two competing transfer effects:
\textit{Negative transfer} occurs when the gradient signal of weakly related data distributions across modalities is conflicting during joint optimization, hindering modality-specific feature learning \cite{jiang2023forkmerge,wang2019characterizing}.
\textit{Positive transfer} arises when shared (anatomy, intensity etc.) structures across distributions provide a complementary signal, potentially benefiting low-prevalence classes with limited in-domain data \cite{pos_verma2023longtailchest,pos_huang2023_mtl}.

\textit{Explicit routing} uses a metadata-trained assignment to route data to a (modality-)specific expert. 
Focusing on a single modality prevents negative transfer but limits positive transfer.
\textit{Implicit routing} learns assignments end-to-end, enabling transfer learning.
Yet without careful initialization, implicit routing collapses to a single expert, yielding uninterpretable, inefficient routing while reliable metadata goes unused.
Neither approach simultaneously ensures interpretable routing, cross-modal feature sharing, and prevents negative transfer effects, a limitation \textit{GS-MoE} resolves.

\subsection{Generalist-Specialist-MoE (GS-MoE)}
\label{sec:methods_gs_moe}

\begin{figure}[t]
    \centering
    \includegraphics[width=0.8\linewidth]{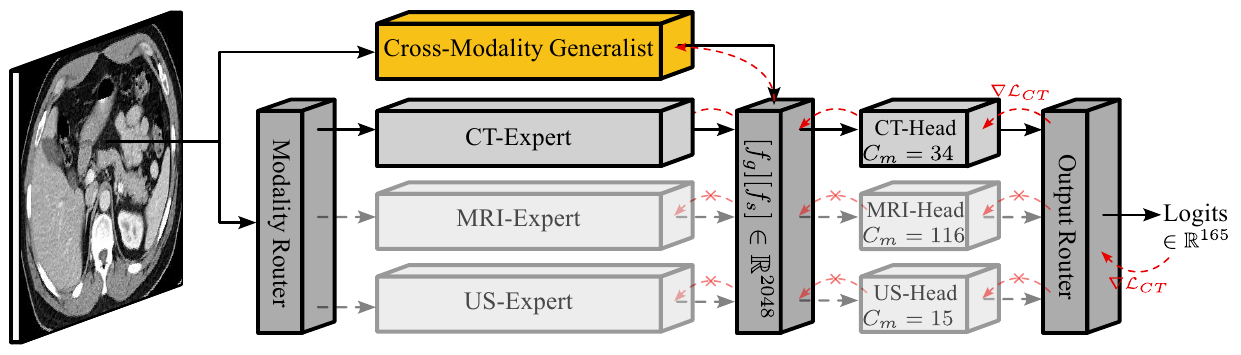}
    \caption{Schematic representation of GS-MoE with the Expert Classifier. Showing an
    exemplary forward (black) and backward (red) pass for a CT input. Training is performed in stages, with each part trained on its relevant subset of data.}
    \label{fig:gs_moe_arch}
\end{figure}

To resolve this trade-off, we propose \textit{Generalist-Specialist-MoE} \textit{(GS-MoE)}, schematized in \autoref{fig:gs_moe_arch}: a two-branch architecture coupling a cross-modal generalist with modality-specific experts that reduce exposure to cross-domain negative transfer.
The \textit{Generalist Branch} acts as an always-active shared pathway, analogous to shared-expert isolation in \ac{MoE} architectures~\cite{dai2024deepseekmoe}, but realized as a full-capacity branch pre-trained across all modalities that also seeds the specialists, learning a shared representation $\mathbf{f}_g \in \mathbb{R}^{d}$ benefiting from positive transfer. The \textit{Specialist Branch} is an \ac{MoE} with explicit routing: a frozen metadata-trained router identifies modality $m$ and activates the relevant expert, yielding domain-constrained features $\mathbf{f}_s \in \mathbb{R}^{d}$ while keeping routing interpretable by construction.
The branches are fused by concatenation, $\mathbf{f}_\text{fused} = [\mathbf{f}_g \| \mathbf{f}_s] \in \mathbb{R}^{2d}$, and jointly fine-tuned with the router frozen.
To enforce complementary representations, we apply a cosine orthogonality penalty inspired by~\cite{bousmalis2016domain}, as $\mathcal{L}_\text{ortho} = B^{-1} \sum_i (\mathbf{f}_g^\top \mathbf{f}_s \|\mathbf{f}_g\|^{-1} \|\mathbf{f}_s\|^{-1})^2$, with $B$ the batch size.

\subsection{Domain-Constrained Optimization: Why Shared Heads Fail}
\label{sec:methods_gradient_dilution}

$\mathbf{f}_\text{fused}$ exposes a structural vulnerability when passed through a shared output head.
Consider $D$ domains with $|C_m|$ classes in domain $m$ and $|C_\text{total}| = \sum_m |C_m|$.
A shared head $\phi_{\text{shared}}\colon \mathbb{R}^{2d} \rightarrow \mathbb{R}^{|C_\text{total}|}$ applied to a domain-$m$ sample computes the loss over all $|C_\text{total}|$, including clinically impossible ones.
Gradients then flow through all $|C_\text{total}|$ outputs, with the $|C_\text{total}| - |C_m|$ non-domain outputs diluting the signal.
We address this via modality-specific network heads $\{\phi_m\}$, each mapping $\mathbb{R}^{2d} \rightarrow \mathbb{R}^{|C_m|}$.
While restricting predictions to the modality's valid-label subset is an established masking operation \cite{cao2019}, our contribution is the finding that its benefit is not diffuse but concentrated on low-prevalence classes, consistent with removal of cross-domain gradient interference where in-domain data is scarcest.
We ablate this in \autoref{tab:experiments_ablation_compact} and find that this raises MR-class macro-F1 from $0.4092$ to $0.4537$ ($+0.045$) on low-prevalence classes.
During training, metadata selects the correct head, inducing $\mathcal{L}_{\text{CE}_m} = -\sum_{c \in C_m} y_c \log \hat{p}_c$, with $\mathcal{L} = \mathcal{L}_{\text{CE}_m} + \lambda\,\mathcal{L}_\text{ortho}$. \\
At inference, $\phi_m$ produces $|C_m|$ logits, which a fixed, expert-knowledge-driven assignment matrix maps to the global $|C_\text{total}|$-dimensional label space.

\section{Experimental Setup}\label{sec:experiments}

\textbf{Dataset.}\label{sec:experiments_dataset}
We evaluate on RadImageNet~\cite{mei2022radimagenet}, a large multimodal pathology classification dataset with 1.35M 2D images across \ac{CT}, \ac{MR}, and \ac{US}, comprising 165 classes: $|C_\text{CT}|=34$, $|C_\text{MR}|=116$, $|C_\text{US}|=15$.  
All models use the official train/val/test splits.
\textbf{Generalist-Specialist-MoE (GS-MoE)}\label{sec:experiments_hybrid_explicit_moe}  
We implement GS-MoE with\linebreak\ Dense\-Net\-121 (pre-classifier output dimensionality $d=1024$) for both generalist and specialists, yielding $\mathbf{f}_\text{fused} \in \mathbb{R}^{2048}$ with domain-specific expert classifiers \linebreak\mbox{$\phi_m\colon 2048 \rightarrow 1024 \rightarrow |C_m|$} and modality router, a lightweight DenseNet29 (4 blocks, $\text{growth rate}=8$, $\sim$61k params, $>99.9\%$ modality prediction accuracy). 
Routing is modality-based, which outperforms modality × anatomy experts.
Training follows a three-stage curriculum (ablated in \autoref{tab:experiments_results_main}): (1) generalist pre-training on all 165 classes, (2) specialist generation via sparse upcycling~\cite{komatsuzaki2023sparseupcycling} from the converged generalist and modality-specific fine-tuning, and (3) GS-MoE construction and joint fine-tuning of branches and classifiers.  
This procedure preserves cross-modal features while enabling high-capacity, low-cost training.
\textbf{Baselines.}\label{sec:experiments_baselines}  
As commonly used in medical imaging classification tasks \cite{chambon2024chexpertplusaugmentinglarge,santamato2025multilabel}, we consider \textit{DenseNet121} ($7.1$M), \textit{DenseNet161} ($26.8$M)~\cite{huang2018densenet}, and a parameter-matched \textit{DenseNet161Ext} (blocks: 6/12/36/38, $35.0$M) trained on the full corpus as dense monolithic baselines.  
As \ac{MoE} baselines, we use  
\textit{ExplicitMoE}, identical to GS-MoE but without the generalist branch, and  
\textit{ImplicitMoE}, which shares the architecture yet is trained end-to-end.  
\textbf{Remark:} While we use DenseNets due to slightly higher performance, our findings directly translate to ViT-T/ViT-S backbones, indicating the effect is architectural (see \autoref{sec:discussion}).
\textbf{Training Setup and Metrics.}\label{sec:experiments_training}
Models are trained until convergence with AdamW, cosine annealing ($T_0=4$), on $224{\times}224$ inputs with random affine and horizontal flips.  
Hyperparameters are empirically optimized. 
We report \ac{MCC} as the primary metric for its robustness to class imbalance, alongside macro-F1.

\section{Results}

\subsection{Aggregate Results} 

\begin{figure}[bhtp]
    \centering
    \resizebox{0.8\textwidth}{!}{
        \input{figures/pareto_scatter}
    }
    \caption{MCC vs.\ active parameters. Dashed line: Pareto frontier.  
    GS-MoE (Expert Classifier + $L_{ortho}$) is Pareto optimal w.r.t. parameter efficiency, outperforming explicitly and implicitly routed \ac{MoE} as well as dense models. 
    }
    \label{fig:pareto}
\end{figure}
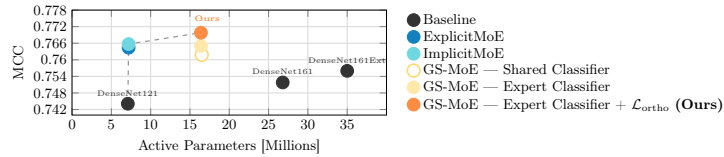

\autoref{tab:experiments_results_main} summarizes test-set performance across all architectures. Without sparse upcycling MoEs collapse to ExplicitMoE MCC of $0.639$ and ImplicitMoE MCC of $0.691$. Mean and standard deviation are computed across 9 seeds throughout.
The ranking in terms of MCC across DenseNet161 < ExplicitMoE < GS-MoE (Expert) is statistically significant ($p<0.05$, paired t-test).
At inference, GS-MoE activates only the generalist, the frozen router, and one expert (${\sim}16.4$M parameters), using ${\sim}53\%$ fewer active parameters than DenseNet161Ext ($35.0$M parameters).  
This reduction is structurally guaranteed by routing and is Pareto-optimal in terms of parameter efficiency (see \autoref{fig:pareto}).
\begin{table}[htbp]
\footnotesize
    \centering
    \setlength{\aboverulesep}{0pt}
    \setlength{\belowrulesep}{0pt}

    \caption{\acp{MoE} outperform dense models. GS-MoE achieves top MCC/F1 with ${\sim}53\%$ fewer active params. than DenseNet161Ext. \textit{Best in family}, \textbf{Best overall}.}
    \label{tab:experiments_results_main}

    \begin{NiceTabular}{llcccc}[]
    \toprule
    Family & Model & MCC & macro-F1 & Active Params  & Total Params\\
    \midrule
    \Block{2-1}{Baseline}
        & DenseNet121 & 0.744$^{\pm0.022}$ & 0.751$^{\pm0.020}$ &  7.1M & 7.1M\\
        & DenseNet161 & 0.752$^{\pm0.018}$ & 0.759$^{\pm0.016}$ &  26.8M & 26.8M\\
        & DenseNet161Ext & \textit{0.756}$^{\pm0.012}$ & \textit{0.763}$^{\pm0.010}$ &  35.0M & 35.0M\\
    \midrule
    \Block{2-1}{\ac{MoE}}
        & Explicit & 0.764$^{\pm0.003}$ & 0.771$^{\pm0.004}$ & 7.2M & 21.4M \\
        & Implicit & \textit{0.766}$^{\pm0.005}$ & \textit{0.772}$^{\pm0.006}$ & 7.2M & 21.4M \\
    \midrule

    \rowcolor{reallylightgray}
    \Block{2-1}{%
        \textcolor{GSMoEOrange}{GS-MoE} \\
    } & Shared Classifier & 0.762$^{\pm0.005}$ & 0.768$^{\pm0.005}$ & 16.5M & 30.8M\\

    \rowcolor{reallylightgray}
    & Expert Classifier & \textbf{0.770}$^{\pm0.003}$ & \textbf{0.776}$^{\pm0.004}$ & 16.4M & 35.0M\\
    \bottomrule
    \end{NiceTabular}
\end{table}

\subsection{Routing Strategies Face Mutually Exclusive Failure Modes}\label{sec:experiments_routing_dilemma}

Our evaluations show that monolithic models and baseline \acp{MoE} face mutually exclusive failure modes, motivating the proposed generalist-specialist \ac{MoE}.

\textbf{Negative transfer and pre-training.} \label{sec:experiments_negative_transfer}
Scaling monolithic capacity shows diminishing returns: DenseNet161Ext improves over DenseNet121 by only $+1.2$\,pp in MCC at $4.9\times$ more parameters, indicating an architectural ceiling.
GS-MoE resolves this limit and outperforms DenseNet121 by $+2.6$\,pp, with matched total and ${\sim}53\%$ fewer active parameters.

\textbf{Cost of explicit routing on MR domains.}\label{sec:experiments_divide_penalty}
Explicit routing improves visually isolated domains (CT-Lung $+0.041$, US $+0.008$ group/\allowbreak modality aggregated F1) but degrades all eight MR groups (e.g., MR-Brain $-0.087$, MR-Abdomen $-0.064$, MR-Hip $-0.051$, MR-Ankle $-0.035$).  
At the pathology level, as evidenced in \autoref{fig:low_preval}, ExplicitMoE improves on none of the 20\% lowest-performing classes relative to DenseNet161, with brain-pituitary collapsing $0.727 \rightarrow 0.000$, AF-coalition $0.400 \rightarrow 0.000$, and AF-spring ligament $0.316 \rightarrow 0.000$, indicating that ExplicitMoE may not effectively leverage cross-domain information for these classes.

\textbf{Implicit routing recovers representation but not semantics.} \label{sec:experiments_implicit_recovery}
Without pre-training, ImplicitMoE deposits $93.6\%$ of inputs to a single expert despite an auxiliary load-balancing loss~\cite{shazeer2017outrageouslylargenn}.  
Sparse upcycling restores a distributed routing profile ($41.5/34.5/24.0\%$).  
Clustering analysis reveals that the router forms data-driven groups: one expert specializes in CT-Lung imagery ($91.1\%$ CT, $90.5\%$ lung), while the remaining experts process a heterogeneous mix.  
Despite this, as the only baseline model, ImplicitMoE outperforms DenseNet161 across all 11 anatomical groups, with the largest gains on MR-Ankle ($+0.083$), MR-Spine ($+0.050$). 
Nonetheless, lacking explicit global cross-modal features, both routing strategies fail on the low-prevalence classes (\autoref{fig:low_preval}).

\begin{figure}[htbp]
    \centering
    \begin{minipage}{0.49\textwidth}
        \centering
        \resizebox{\textwidth}{!}{
            \input{figures/low_preval.tex}
        }
        \captionof{figure}{
    Sliding-window (width=20) macro F1 (90\% CI) over prevalence-ordered classes. ExplicitMoE trails all baselines on low-prevalence classes due to a lack of cross-domain information; GS-MoE recovers this regime, matching or improving throughout. 
        }
        \label{fig:low_preval}
    \end{minipage}
    \hfill
\begin{minipage}{0.49\textwidth}
    \centering
    \footnotesize
    \renewcommand{\arraystretch}{1.1}
    \setlength{\tabcolsep}{3pt}
    \setlength{\aboverulesep}{0pt}
    \setlength{\belowrulesep}{0pt}
    \captionof{table}{
        Top per-class gains/losses of GS-MoE vs.\ per-class best baseline.
        Classes marked with $^*$ have F1 $=$ 0 across all baselines and recover through cross-modal information.
        AF: Ankle/Foot, Abd: Abdomen, Sp: Spine.
    }
    \label{tab:experiments_results_gains}
\begin{NiceTabular}{@{}l@{\hspace{\dimexpr 2\tabcolsep-5mm\relax}}c@{\hspace{\dimexpr 2\tabcolsep-2mm\relax}}r@{}}[]
    \toprule
    Category & \counts{$N_{\text{train}}$}{$N_{\text{test}}$} & F1 $\Delta$ \\
    \midrule
    Lisfranc$_{\,\text{MR-AF}}$          & \counts{37}{8}     & \color{Endeavour} $+$0.60$^{\pm0.12}$$^{*}$ \\
    Spring Lig.$_{\,\text{MR-AF}}$       & \counts{83}{5}     & \color{Endeavour} $+$0.56$^{\pm0.20}$$^{*}$ \\
    Coalition$_{\,\text{MR-AF}}$         & \counts{50}{4}     & \color{Endeavour} $+$0.56$^{\pm0.36}$$^{*}$ \\
    Hematoma$_{\,\text{MR-AF}}$          & \counts{165}{29}   & \color{Endeavour} $+$0.55$^{\pm0.04}$$^{\,\,\,}$ \\
    Extensor$_{\,\text{MR-AF}}$          & \counts{536}{83}   & \color{Endeavour} $+$0.44$^{\pm0.03}$$^{\,\,\,}$ \\
    Intra-Art. Mass$_{\,\text{MR-AF}}$   & \counts{455}{60}   & \color{Endeavour} $+$0.40$^{\pm0.03}$$^{\,\,\,}$ \\
    Neoplasm$_{\,\text{MR-AF}}$          & \counts{745}{95}    & \color{Endeavour} $+$0.33$^{\pm0.02}$$^{\,\,\,}$ \\ 
    Prostate$_{\,\text{MR-Abd}}$         & \counts{53}{4}     & \color{Endeavour} $+$0.24$^{\pm0.26}$$^{\,\,\,}$ \\ \midrule
    Prostate Les.$_{\,\text{CT-Abd}}$ & \counts{782}{98}   & \color{GSMoEOrange} $-$0.05$^{\pm0.02}$$^{\,\,\,}$ \\
    Bladder$_{\,\text{CT-Abd}}$          & \counts{1477}{157}   & \color{GSMoEOrange} $-$0.07$^{\pm0.02}$$^{\,\,\,}$ \\
    Renal Les.$_{\,\text{MR-Abd}}$     & \counts{1627}{234}   & \color{GSMoEOrange} $-$0.10$^{\pm0.02}$$^{\,\,\,}$ \\
    Normal CT$_{\,\text{CT-Abd}}$        & \counts{57432}{6689} & \color{GSMoEOrange} $-$0.11$^{\pm0.03}$$^{\,\,\,}$ \\
    \bottomrule
\end{NiceTabular}
    \end{minipage}
\end{figure}

\begin{figure}[t]
    \centering
    \resizebox{\textwidth}{!}{
        \input{figures/scatter}
    }
    \caption{
    Per-pathology F1-score comparison (165 classes).
    \textbf{(a)} ExplicitMoE improves on distinct modalities (US, CT-Lung) but collapses on MR-AF/Brain, with several classes reaching zero.
    \textbf{(b)} GS-MoE (Expert Classifier + $L_{ortho}$) recovers the MR-AF cluster, outperforming the DenseNet161, including several classes at $x{=}0$ (zero-baseline performance).
    \textbf{(c)} GS-MoE (Expert Classifier + $L_{ortho}$) strongly exceeds ExplicitMoE, particularly in MRI, attributed to explicit generalist-driven cross-domain knowledge.
}
    \label{fig:experiments_results_combined}
\end{figure}
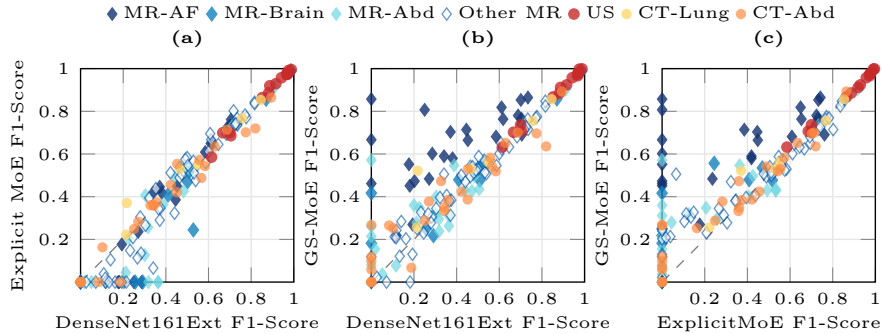

\subsection{Hard Class Recovery: A Qualitative Capability Gap}
\label{sec:experiments_hard_class_recovery}
While ExplicitMoE mitigates negative transfer, gaining slightly on distinct modalities (US, CT-Lung), it lacks positive transfer during fine-tuning, degrading cross-modal knowledge acquired during pre-training.
As illustrated in \autoref{fig:experiments_results_combined} (a), this leads to performance collapse (F1 $=$ 0) on some MR-pathologies.
In contrast, GS-MoE architecturally augments experts with cross-modal features from a dedicated generalist branch.
GS-MoE recovers performance on collapsed pathologies, particularly within the MR-Ankle/Foot group (\autoref{fig:experiments_results_combined} (b, c)), and significantly ($p<0.001$, paired t-test) outperforms all baselines on low-prevalence classes (\autoref{fig:low_preval}), with the strongest recovery in the MR-Abdomen group ($+0.057$), the weakest-performing group overall.
Crucially, GS-MoE achieves non-zero F1 on six low-prevalence pathologies (< 100 samples) where \textbf{all baselines}, including DenseNet161Ext trained with focal loss ($\gamma \in \{1, 2, 5\}$) and post-hoc classifier rebalancing ($\tau$-normalization, $\tau \in [0, 1]$)~\cite{Kang2020Decoupling}, fail entirely (F1 $=$ 0), with recoveries as large as +0.60 (lisfranc pathology), +0.56 (spring ligament) and +0.56 (coalition) (\hyperref[tab:experiments_results_gains]{Table~\ref*{tab:experiments_results_gains}}).
These gains are consistent with the generalist branch providing expert-complementary information.
This is evidenced by GS-MoE outperforming the explicit specialist-only \ac{MoE} across most pathologies (\autoref{fig:experiments_results_combined} (c)), while only differing by the generalist.
As the parameter-matched DenseNet161Ext does not recover the tail (\autoref{fig:low_preval}) despite equal capacity, we attribute the low-prevalence recovery to the generalist branch.
The recovered signal is cross-modal by nature, a capability unimodal experts cannot provide.
Decoupling generalist (cross-modal) and specialist (unimodal) knowledge may yield a more effective feature representation than specialist-only \acp{MoE} and monolithic models.

\subsection{Ablation}\label{sec:experiments_gradient_dilution_ablation}
\autoref{tab:experiments_ablation_compact} ablates the contribution of each part of GS-MoE and finds that \textit{Generalist}, \textit{Specialist}, \textit{Domain Head}, and \textit{$\mathcal{L}_{\text{ortho}}$} all improve MCC.
The gap between the Shared Classifier and Expert Classifier isolates the effect of the shared output head (\autoref{sec:methods_gradient_dilution}), as they are identical except for the output head (\autoref{tab:experiments_ablation_compact}). 
The shared head degrades disproportionately on low-prevalence MR classes: with all else held fixed, switching to domain-constrained heads raises the MR-class-averaged F1 from $0.4092$ to $0.4537$ ($+0.045$,$ ~11\%$ relative).
Moreover, the gain from domain-specific heads points to a benefit from modality-specific processing of cross-domain information. 

\begin{table}[t]
    \centering
    \footnotesize
    \setlength{\tabcolsep}{4pt}
    \caption{Ablation of Domain-Constrained Optimization. The gap between the Shared Classifier and Expert Classifier isolates the effect of the expert head. 
    DenseNet161Ext and ExplicitMoE serve as reference points. 
    }
    \label{tab:experiments_ablation_compact}
    \begin{NiceTabular}{l @{\hspace{5pt}} ccccc}
        \toprule
        Model Variant & Generalist & Specialist & Domain Head & $\mathcal{L}_{\text{ortho}}$ & MCC \\
        \midrule
        DenseNet161Ext & \cmark & \xmark & \xmark & \xmark & 0.756 \\
        ExplicitMoE & \xmark & \cmark & \cmark & \xmark & 0.764 \\
        \midrule
        GS-MoE \scriptsize{(Shared Classifier)} & \cmark & \cmark & \xmark & \xmark & 0.762 \\
        GS-MoE \scriptsize{(Expert Classifier)} & \cmark & \cmark & \cmark & \xmark & 0.765 \\
        GS-MoE \scriptsize{(Expert C. + $\mathcal{L}_\text{ortho}$)} & \cmark & \cmark & \cmark & \cmark & \textbf{0.770} \\
        \bottomrule
    \end{NiceTabular}
\end{table}

\section{Discussion}
\label{sec:discussion}
GS-MoE mitigates the fundamental conflict between modality-specific specialization and cross-modal sharing: low-prevalence collapse under modality routing is consistent with cross-domain gradient interference, recoverable via an always-active generalist. 
Coupling a modality-routed specialist MoE with this generalist preserves interpretable routing while retaining shared cross-modal representations.
We show that GS-MoE outperforms all investigated baselines, with the margin most pronounced in low-prevalence classes.
While we showed results on dense backbones, we found the same trends with transformer backbones, with MCC of: ViT-T/S baselines 0.726/0.737, Ex\-pli\-cit\-MoE-ViT-T 0.741, GS-MoE-ViT-T 0.745. 
Although absolute MCC is lower than for DenseNet, the ordering (GS-MoE > ExplicitMoE > Dense) and the low-prevalence recovery are fully preserved, indicating the generalist-specialist effect is architectural rather than backbone-specific.
The limitations of our work directly motivate future work extending GS-MoE to (volumetric) segmentation, integrating report-grounded supervision, developing hierarchical experts conditioned on anatomy and demographics, potentially as hierarchical \acp{MoE}, and ablations across different model sizes.
In summary, GS-MoE shows that metadata-supervised specialization and learned generalist features contribute complementary gains, a practical step toward rare-pathology-aware multimodal imaging.

\bibliographystyle{splncs04}
\bibliography{main}
\end{document}

%% file: figures/pareto_scatter.tex
\begin{tikzpicture}
\pgfplotsset{compat=1.18}
\begin{axis}[
    xlabel={Active Parameters [Millions]},
    ylabel={MCC},
    width=0.7\textwidth,
    height=0.32\textwidth,
    grid=major,
    grid style={gray!30, line width=0.5pt},
    tick label style={font=\small},
    label style={font=\small},
    xmin=0,
    xmax=40,
    ymin=0.740,
    ymax=0.778,
    xtick={0,5,10,15,20,25,30, 35},
    ytick={0.742,0.748,0.754,0.760,0.766,0.772,0.778},
    yticklabel style={/pgf/number format/fixed, /pgf/number format/precision=3},
    legend style={
        draw=none,
        at={(1.05, 0.5)},
        anchor=west,
        font=\footnotesize, 
        row sep=-1pt
    },
    legend cell align={left},
]

\addplot[
    TUMDarkGray!50, dashed, line width=0.8pt,
    forget plot,
] coordinates {
    (7.1, 0.7441)
    (7.2, 0.7657)
    (16.4, 0.7698)
};

\addplot[
    only marks, mark=*, mark size=4pt,
    color=TUMDarkGray, mark options={solid},
] coordinates {
    (7.1,  0.7441)   
    (26.8, 0.7518)   
    (35.0, 0.756) 
};
\addlegendentry{Baseline}

\addplot[
    only marks, mark=*, mark size=4pt,
    color=Horizon,
] coordinates {
    (7.2, 0.7644)
};
\addlegendentry{ExplicitMoE}

\addplot[
    only marks, mark=*, mark size=4pt,
    color=ModelUltramarine,
] coordinates {
    (7.2, 0.7657)
};
\addlegendentry{ImplicitMoE}

\addplot[
    only marks, 
    mark=*, 
    mark size=4pt,
    mark options={
        line width=1pt,
        draw=GSMoEYellow, 
        fill=white        
    }
] coordinates {
    (16.5, 0.7618)   
};
\addlegendentry{GS-MoE — Shared Classifier}

\addplot[
    only marks, mark=*, mark size=4pt,
    color=GSMoEYellow!60!white,
] coordinates {
    (16.4, 0.7650)   
};
\addlegendentry{GS-MoE — Expert Classifier}

\addplot[
    only marks, mark=*, mark size=4pt,
    color=GSMoEOrange,
] coordinates {
    (16.4, 0.7698)   
};
\addlegendentry{GS-MoE — Expert Classifier + $\mathcal{L}_\text{ortho}$ \textbf{(Ours)}}


\node[above, font=\tiny, color=TUMDarkGray]
    at (axis cs: 7.1, 0.7450) {DenseNet121};
\node[above, font=\tiny, color=TUMDarkGray]
    at (axis cs: 26.8, 0.7528) {DenseNet161};
\node[above, font=\tiny, color=TUMDarkGray]
    at (axis cs: 35.0, 0.757) {DenseNet161Ext};
\node[above left, font=\tiny, color=PlotDeepBlue]
    at (axis cs: 7.2, 0.7644) {};
\node[below right, font=\tiny, color=PlotSkyBlue]
    at (axis cs: 7.2, 0.7657) {};
\node[above, font=\tiny, color=GSMoEOrange]
    at (axis cs: 17.2, 0.772) {\textbf{Ours}};

\end{axis}
\end{tikzpicture}

%% file: figures/low_preval.tex
\begin{tikzpicture}[font=\tiny]
\begin{axis}[
    width=3.4cm, height=2.9cm, scale only axis,
    xmode=log, log basis x=10,
    xlabel={$N_{train}$ per class}, ylabel={Macro F1},
    xmin=48.4, xmax=52139.9,
    ymin=-0.02, ymax=0.90, ytick={0,0.2,0.4,0.6,0.8},
    xtick={100,1000,10000}, xticklabels={$10^2$,$10^3$,$10^4$},
    grid=both, minor grid style={gray!10}, major grid style={gray!20},
    tick align=outside, tick pos=left,
    label style={font=\tiny}, tick label style={font=\tiny},
    x label style={yshift=2pt}, y label style={yshift=-3pt},
    every tick/.style={gray, line width=0.5pt},
    every axis plot/.append style={line width=0.8pt, mark=none},
    legend style={font=\tiny, draw=none, fill=white, fill opacity=0.6, text opacity=1, inner sep=1pt, row sep=-1pt, nodes={inner sep=1pt}},
    legend image post style={line width=1pt},
    legend cell align=left,
    legend pos=north west,
]
\addplot[draw=none, forget plot, name path=U0] coordinates {(77.5,0.0812) (83.0,0.0853) (88.8,0.0971) (94.2,0.1013) (100.0,0.1113) (106.1,0.1072) (112.2,0.1115) (118.6,0.1141) (125.0,0.1150) (131.6,0.1156) (138.5,0.1300) (145.8,0.1302) (153.0,0.1466) (160.1,0.1513) (167.0,0.1532) (176.5,0.1590) (185.7,0.1592) (194.8,0.1563) (204.1,0.1666) (214.4,0.1689) (225.5,0.1650) (236.6,0.1626) (248.2,0.1785) (260.0,0.1759) (272.1,0.2051) (284.3,0.2192) (296.7,0.2350) (309.6,0.2313) (325.2,0.2341) (341.5,0.2461) (357.8,0.2457) (374.9,0.2514) (393.8,0.2596) (412.9,0.2567) (432.8,0.2808) (447.6,0.2848) (462.3,0.3076) (477.4,0.3354) (492.2,0.3672) (506.1,0.3652) (519.3,0.3654) (535.2,0.3883) (551.0,0.3745) (566.4,0.3807) (582.7,0.3590) (600.8,0.3886) (617.9,0.3772) (635.4,0.4076) (650.2,0.4286) (667.0,0.4397) (684.0,0.4461) (702.0,0.4439) (722.1,0.4505) (744.5,0.4642) (767.9,0.4634) (792.2,0.4598) (817.9,0.4514) (844.5,0.4423) (872.7,0.4164) (903.3,0.4274) (936.5,0.4287) (968.4,0.4173) (1001.3,0.4205) (1035.2,0.4352) (1070.3,0.4516) (1104.6,0.4205) (1139.6,0.4325) (1176.3,0.4237) (1212.7,0.4129) (1249.5,0.4039) (1288.0,0.4130) (1327.6,0.4502) (1362.9,0.4608) (1397.2,0.4551) (1431.3,0.4429) (1468.4,0.4529) (1507.1,0.4650) (1546.5,0.4719) (1585.3,0.4725) (1622.6,0.4682) (1661.7,0.5109) (1699.3,0.5148) (1737.6,0.5517) (1777.0,0.5460) (1817.0,0.5500) (1861.3,0.5490) (1915.1,0.5518) (1980.0,0.5451) (2046.1,0.5655) (2113.8,0.5711) (2180.8,0.5918) (2248.9,0.6021) (2316.7,0.5935) (2394.5,0.6084) (2477.3,0.6118) (2569.4,0.6242) (2663.6,0.6168) (2769.6,0.6131) (2892.7,0.6286) (3031.4,0.6481) (3176.3,0.6487) (3334.4,0.6509) (3502.5,0.6303) (3682.4,0.6306) (3871.3,0.6288) (4062.4,0.6314) (4249.6,0.6383) (4426.7,0.6428) (4612.5,0.6247) (4801.2,0.6376) (5014.4,0.6220) (5241.1,0.6068) (5487.3,0.6101) (5723.0,0.6033) (5973.1,0.6329) (6205.1,0.6377) (6451.2,0.6536) (6691.7,0.6735) (6913.7,0.6762) (7131.9,0.6681) (7341.9,0.6456) (7539.6,0.6593) (7737.3,0.6552) (7970.4,0.6853) (8331.0,0.7126) (8735.3,0.7291) (9161.4,0.7369) (9630.0,0.7623) (10193.1,0.7718) (10789.7,0.7655) (11435.7,0.7694) (12119.3,0.7884) (12822.4,0.7903) (13592.2,0.7928) (14414.0,0.7856) (15299.8,0.7868) (16244.9,0.8052) (17303.7,0.8060) (18451.8,0.8229) (19718.8,0.8294) (21470.3,0.8366) (23401.6,0.8523) (25556.4,0.8611) (27814.0,0.8603) (29867.8,0.8595) (32587.4,0.8716)};
\addplot[draw=none, forget plot, name path=L0] coordinates {(77.5,0.0120) (83.0,0.0164) (88.8,0.0244) (94.2,0.0292) (100.0,0.0376) (106.1,0.0345) (112.2,0.0399) (118.6,0.0411) (125.0,0.0427) (131.6,0.0438) (138.5,0.0542) (145.8,0.0546) (153.0,0.0663) (160.1,0.0735) (167.0,0.0762) (176.5,0.0847) (185.7,0.0851) (194.8,0.0840) (204.1,0.0958) (214.4,0.0993) (225.5,0.0986) (236.6,0.0930) (248.2,0.0962) (260.0,0.0912) (272.1,0.0965) (284.3,0.1076) (296.7,0.1188) (309.6,0.1143) (325.2,0.1195) (341.5,0.1340) (357.8,0.1337) (374.9,0.1443) (393.8,0.1479) (412.9,0.1416) (432.8,0.1572) (447.6,0.1619) (462.3,0.1848) (477.4,0.1979) (492.2,0.2124) (506.1,0.2074) (519.3,0.2076) (535.2,0.2348) (551.0,0.2160) (566.4,0.2297) (582.7,0.2104) (600.8,0.2212) (617.9,0.2071) (635.4,0.2311) (650.2,0.2529) (667.0,0.2625) (684.0,0.2701) (702.0,0.2650) (722.1,0.2697) (744.5,0.2938) (767.9,0.2932) (792.2,0.2854) (817.9,0.2774) (844.5,0.2729) (872.7,0.2572) (903.3,0.2775) (936.5,0.2792) (968.4,0.2702) (1001.3,0.2796) (1035.2,0.2940) (1070.3,0.3161) (1104.6,0.2985) (1139.6,0.3187) (1176.3,0.3166) (1212.7,0.3099) (1249.5,0.3035) (1288.0,0.3085) (1327.6,0.3368) (1362.9,0.3403) (1397.2,0.3324) (1431.3,0.3165) (1468.4,0.3358) (1507.1,0.3469) (1546.5,0.3506) (1585.3,0.3519) (1622.6,0.3439) (1661.7,0.3622) (1699.3,0.3680) (1737.6,0.3967) (1777.0,0.3903) (1817.0,0.3933) (1861.3,0.3904) (1915.1,0.3933) (1980.0,0.3866) (2046.1,0.4024) (2113.8,0.4102) (2180.8,0.4205) (2248.9,0.4234) (2316.7,0.4167) (2394.5,0.4362) (2477.3,0.4461) (2569.4,0.4585) (2663.6,0.4480) (2769.6,0.4445) (2892.7,0.4649) (3031.4,0.4927) (3176.3,0.4928) (3334.4,0.4968) (3502.5,0.4842) (3682.4,0.4849) (3871.3,0.4831) (4062.4,0.4946) (4249.6,0.5017) (4426.7,0.5094) (4612.5,0.4864) (4801.2,0.4978) (5014.4,0.4905) (5241.1,0.4866) (5487.3,0.4901) (5723.0,0.4833) (5973.1,0.5075) (6205.1,0.5100) (6451.2,0.5303) (6691.7,0.5440) (6913.7,0.5460) (7131.9,0.5388) (7341.9,0.5286) (7539.6,0.5421) (7737.3,0.5326) (7970.4,0.5556) (8331.0,0.5710) (8735.3,0.5969) (9161.4,0.6035) (9630.0,0.6225) (10193.1,0.6505) (10789.7,0.6412) (11435.7,0.6464) (12119.3,0.6565) (12822.4,0.6601) (13592.2,0.6674) (14414.0,0.6626) (15299.8,0.6636) (16244.9,0.6734) (17303.7,0.6739) (18451.8,0.6852) (19718.8,0.6965) (21470.3,0.7080) (23401.6,0.7184) (25556.4,0.7496) (27814.0,0.7491) (29867.8,0.7488) (32587.4,0.7573)};
\addplot[TUMDarkGray, fill opacity=0.18, forget plot] fill between[of=U0 and L0];
\addplot[draw=none, forget plot, name path=U1] coordinates {(77.5,0.0000) (83.0,0.0000) (88.8,0.0000) (94.2,0.0000) (100.0,0.0000) (106.1,0.0000) (112.2,0.0000) (118.6,0.0000) (125.0,0.0000) (131.6,0.0000) (138.5,0.0000) (145.8,0.0000) (153.0,0.0000) (160.1,0.0000) (167.0,0.0000) (176.5,0.0000) (185.7,0.0000) (194.8,0.0000) (204.1,0.0053) (214.4,0.0053) (225.5,0.0091) (236.6,0.0091) (248.2,0.0147) (260.0,0.0147) (272.1,0.0779) (284.3,0.0839) (296.7,0.1004) (309.6,0.1004) (325.2,0.1004) (341.5,0.1069) (357.8,0.1164) (374.9,0.1164) (393.8,0.1353) (412.9,0.1353) (432.8,0.1720) (447.6,0.1720) (462.3,0.1930) (477.4,0.2320) (492.2,0.2775) (506.1,0.2775) (519.3,0.2846) (535.2,0.3096) (551.0,0.3074) (566.4,0.3157) (582.7,0.2878) (600.8,0.3269) (617.9,0.3161) (635.4,0.3573) (650.2,0.3849) (667.0,0.4037) (684.0,0.4151) (702.0,0.4203) (722.1,0.4175) (744.5,0.4305) (767.9,0.4333) (792.2,0.4337) (817.9,0.4335) (844.5,0.4302) (872.7,0.4035) (903.3,0.4163) (936.5,0.4199) (968.4,0.4132) (1001.3,0.4197) (1035.2,0.4326) (1070.3,0.4568) (1104.6,0.4254) (1139.6,0.4391) (1176.3,0.4257) (1212.7,0.4145) (1249.5,0.4064) (1288.0,0.4148) (1327.6,0.4555) (1362.9,0.4809) (1397.2,0.4745) (1431.3,0.4590) (1468.4,0.4705) (1507.1,0.4786) (1546.5,0.4855) (1585.3,0.4869) (1622.6,0.4830) (1661.7,0.5208) (1699.3,0.5242) (1737.6,0.5596) (1777.0,0.5557) (1817.0,0.5555) (1861.3,0.5550) (1915.1,0.5552) (1980.0,0.5509) (2046.1,0.5715) (2113.8,0.5773) (2180.8,0.5997) (2248.9,0.6063) (2316.7,0.5969) (2394.5,0.6168) (2477.3,0.6172) (2569.4,0.6320) (2663.6,0.6263) (2769.6,0.6194) (2892.7,0.6339) (3031.4,0.6552) (3176.3,0.6596) (3334.4,0.6603) (3502.5,0.6429) (3682.4,0.6451) (3871.3,0.6439) (4062.4,0.6494) (4249.6,0.6603) (4426.7,0.6658) (4612.5,0.6499) (4801.2,0.6620) (5014.4,0.6449) (5241.1,0.6326) (5487.3,0.6335) (5723.0,0.6272) (5973.1,0.6551) (6205.1,0.6577) (6451.2,0.6749) (6691.7,0.6934) (6913.7,0.6973) (7131.9,0.6859) (7341.9,0.6617) (7539.6,0.6753) (7737.3,0.6707) (7970.4,0.6989) (8331.0,0.7260) (8735.3,0.7404) (9161.4,0.7468) (9630.0,0.7703) (10193.1,0.7791) (10789.7,0.7723) (11435.7,0.7772) (12119.3,0.7945) (12822.4,0.7982) (13592.2,0.8005) (14414.0,0.7938) (15299.8,0.7955) (16244.9,0.8112) (17303.7,0.8123) (18451.8,0.8283) (19718.8,0.8362) (21470.3,0.8437) (23401.6,0.8585) (25556.4,0.8642) (27814.0,0.8633) (29867.8,0.8622) (32587.4,0.8743)};
\addplot[draw=none, forget plot, name path=L1] coordinates {(77.5,0.0000) (83.0,0.0000) (88.8,0.0000) (94.2,0.0000) (100.0,0.0000) (106.1,0.0000) (112.2,0.0000) (118.6,0.0000) (125.0,0.0000) (131.6,0.0000) (138.5,0.0000) (145.8,0.0000) (153.0,0.0000) (160.1,0.0000) (167.0,0.0000) (176.5,0.0000) (185.7,0.0000) (194.8,0.0000) (204.1,0.0000) (214.4,0.0000) (225.5,0.0000) (236.6,0.0000) (248.2,0.0002) (260.0,0.0002) (272.1,0.0000) (284.3,0.0000) (296.7,0.0032) (309.6,0.0032) (325.2,0.0032) (341.5,0.0093) (357.8,0.0174) (374.9,0.0174) (393.8,0.0290) (412.9,0.0290) (432.8,0.0436) (447.6,0.0436) (462.3,0.0589) (477.4,0.0785) (492.2,0.0991) (506.1,0.0991) (519.3,0.1082) (535.2,0.1307) (551.0,0.1264) (566.4,0.1390) (582.7,0.1152) (600.8,0.1349) (617.9,0.1201) (635.4,0.1488) (650.2,0.1759) (667.0,0.1936) (684.0,0.2054) (702.0,0.2164) (722.1,0.2132) (744.5,0.2345) (767.9,0.2359) (792.2,0.2372) (817.9,0.2371) (844.5,0.2355) (872.7,0.2205) (903.3,0.2424) (936.5,0.2472) (968.4,0.2424) (1001.3,0.2582) (1035.2,0.2714) (1070.3,0.3047) (1104.6,0.2795) (1139.6,0.3078) (1176.3,0.3046) (1212.7,0.2981) (1249.5,0.2927) (1288.0,0.2970) (1327.6,0.3201) (1362.9,0.3389) (1397.2,0.3257) (1431.3,0.3108) (1468.4,0.3383) (1507.1,0.3453) (1546.5,0.3483) (1585.3,0.3508) (1622.6,0.3437) (1661.7,0.3633) (1699.3,0.3675) (1737.6,0.3942) (1777.0,0.3896) (1817.0,0.3894) (1861.3,0.3869) (1915.1,0.3872) (1980.0,0.3831) (2046.1,0.3987) (2113.8,0.4061) (2180.8,0.4168) (2248.9,0.4186) (2316.7,0.4118) (2394.5,0.4421) (2477.3,0.4434) (2569.4,0.4573) (2663.6,0.4492) (2769.6,0.4432) (2892.7,0.4609) (3031.4,0.4891) (3176.3,0.4903) (3334.4,0.4914) (3502.5,0.4811) (3682.4,0.4851) (3871.3,0.4838) (4062.4,0.5098) (4249.6,0.5229) (4426.7,0.5309) (4612.5,0.5129) (4801.2,0.5241) (5014.4,0.5159) (5241.1,0.5128) (5487.3,0.5139) (5723.0,0.5082) (5973.1,0.5394) (6205.1,0.5408) (6451.2,0.5624) (6691.7,0.5769) (6913.7,0.5800) (7131.9,0.5671) (7341.9,0.5552) (7539.6,0.5702) (7737.3,0.5602) (7970.4,0.5803) (8331.0,0.5956) (8735.3,0.6165) (9161.4,0.6217) (9630.0,0.6371) (10193.1,0.6583) (10789.7,0.6481) (11435.7,0.6543) (12119.3,0.6627) (12822.4,0.6700) (13592.2,0.6756) (14414.0,0.6708) (15299.8,0.6723) (16244.9,0.6800) (17303.7,0.6807) (18451.8,0.6910) (19718.8,0.7079) (21470.3,0.7213) (23401.6,0.7313) (25556.4,0.7557) (27814.0,0.7552) (29867.8,0.7548) (32587.4,0.7633)};
\addplot[Horizon, fill opacity=0.18, forget plot] fill between[of=U1 and L1];
\addplot[draw=none, forget plot, name path=U2] coordinates {(77.5,0.1312) (83.0,0.1346) (88.8,0.1386) (94.2,0.1830) (100.0,0.1930) (106.1,0.1836) (112.2,0.1894) (118.6,0.1947) (125.0,0.1961) (131.6,0.1905) (138.5,0.2023) (145.8,0.2043) (153.0,0.2165) (160.1,0.2298) (167.0,0.2213) (176.5,0.2324) (185.7,0.2331) (194.8,0.2248) (204.1,0.2334) (214.4,0.2363) (225.5,0.2361) (236.6,0.2348) (248.2,0.2563) (260.0,0.2179) (272.1,0.2470) (284.3,0.2670) (296.7,0.2813) (309.6,0.2870) (325.2,0.2926) (341.5,0.3007) (357.8,0.3020) (374.9,0.3131) (393.8,0.3198) (412.9,0.3102) (432.8,0.3329) (447.6,0.3347) (462.3,0.3567) (477.4,0.3787) (492.2,0.4157) (506.1,0.4145) (519.3,0.4174) (535.2,0.4370) (551.0,0.4247) (566.4,0.4320) (582.7,0.4201) (600.8,0.4412) (617.9,0.4310) (635.4,0.4599) (650.2,0.4787) (667.0,0.4901) (684.0,0.4983) (702.0,0.4964) (722.1,0.5080) (744.5,0.5183) (767.9,0.5163) (792.2,0.5131) (817.9,0.5076) (844.5,0.4987) (872.7,0.4710) (903.3,0.4764) (936.5,0.4758) (968.4,0.4650) (1001.3,0.4670) (1035.2,0.4786) (1070.3,0.4883) (1104.6,0.4636) (1139.6,0.4710) (1176.3,0.4553) (1212.7,0.4433) (1249.5,0.4335) (1288.0,0.4360) (1327.6,0.4704) (1362.9,0.4824) (1397.2,0.4801) (1431.3,0.4713) (1468.4,0.4814) (1507.1,0.4900) (1546.5,0.4940) (1585.3,0.4991) (1622.6,0.5004) (1661.7,0.5412) (1699.3,0.5464) (1737.6,0.5818) (1777.0,0.5782) (1817.0,0.5783) (1861.3,0.5767) (1915.1,0.5845) (1980.0,0.5851) (2046.1,0.6030) (2113.8,0.6111) (2180.8,0.6359) (2248.9,0.6442) (2316.7,0.6333) (2394.5,0.6496) (2477.3,0.6497) (2569.4,0.6596) (2663.6,0.6531) (2769.6,0.6497) (2892.7,0.6593) (3031.4,0.6759) (3176.3,0.6773) (3334.4,0.6783) (3502.5,0.6581) (3682.4,0.6623) (3871.3,0.6628) (4062.4,0.6673) (4249.6,0.6724) (4426.7,0.6742) (4612.5,0.6596) (4801.2,0.6709) (5014.4,0.6531) (5241.1,0.6452) (5487.3,0.6453) (5723.0,0.6345) (5973.1,0.6627) (6205.1,0.6672) (6451.2,0.6817) (6691.7,0.6993) (6913.7,0.7016) (7131.9,0.6913) (7341.9,0.6688) (7539.6,0.6854) (7737.3,0.6830) (7970.4,0.7077) (8331.0,0.7338) (8735.3,0.7478) (9161.4,0.7548) (9630.0,0.7784) (10193.1,0.7869) (10789.7,0.7801) (11435.7,0.7852) (12119.3,0.7994) (12822.4,0.8048) (13592.2,0.8079) (14414.0,0.8008) (15299.8,0.8024) (16244.9,0.8197) (17303.7,0.8207) (18451.8,0.8373) (19718.8,0.8441) (21470.3,0.8507) (23401.6,0.8631) (25556.4,0.8667) (27814.0,0.8658) (29867.8,0.8647) (32587.4,0.8762)};
\addplot[draw=none, forget plot, name path=L2] coordinates {(77.5,0.0243) (83.0,0.0286) (88.8,0.0337) (94.2,0.0482) (100.0,0.0590) (106.1,0.0514) (112.2,0.0589) (118.6,0.0661) (125.0,0.0686) (131.6,0.0619) (138.5,0.0743) (145.8,0.0779) (153.0,0.0914) (160.1,0.1061) (167.0,0.0988) (176.5,0.1126) (185.7,0.1143) (194.8,0.1130) (204.1,0.1258) (214.4,0.1290) (225.5,0.1289) (236.6,0.1260) (248.2,0.1398) (260.0,0.1218) (272.1,0.1287) (284.3,0.1435) (296.7,0.1547) (309.6,0.1614) (325.2,0.1714) (341.5,0.1853) (357.8,0.1865) (374.9,0.2021) (393.8,0.2070) (412.9,0.1886) (432.8,0.2058) (447.6,0.2076) (462.3,0.2326) (477.4,0.2389) (492.2,0.2567) (506.1,0.2544) (519.3,0.2579) (535.2,0.2852) (551.0,0.2702) (566.4,0.2884) (582.7,0.2815) (600.8,0.2892) (617.9,0.2740) (635.4,0.2919) (650.2,0.3109) (667.0,0.3236) (684.0,0.3337) (702.0,0.3308) (722.1,0.3401) (744.5,0.3688) (767.9,0.3673) (792.2,0.3590) (817.9,0.3536) (844.5,0.3496) (872.7,0.3348) (903.3,0.3471) (936.5,0.3460) (968.4,0.3344) (1001.3,0.3401) (1035.2,0.3523) (1070.3,0.3599) (1104.6,0.3461) (1139.6,0.3642) (1176.3,0.3610) (1212.7,0.3538) (1249.5,0.3439) (1288.0,0.3453) (1327.6,0.3606) (1362.9,0.3632) (1397.2,0.3601) (1431.3,0.3497) (1468.4,0.3678) (1507.1,0.3749) (1546.5,0.3767) (1585.3,0.3831) (1622.6,0.3852) (1661.7,0.4042) (1699.3,0.4136) (1737.6,0.4402) (1777.0,0.4360) (1817.0,0.4361) (1861.3,0.4318) (1915.1,0.4409) (1980.0,0.4414) (2046.1,0.4572) (2113.8,0.4719) (2180.8,0.4856) (2248.9,0.4880) (2316.7,0.4799) (2394.5,0.4998) (2477.3,0.5004) (2569.4,0.5113) (2663.6,0.5000) (2769.6,0.4965) (2892.7,0.5072) (3031.4,0.5272) (3176.3,0.5275) (3334.4,0.5295) (3502.5,0.5171) (3682.4,0.5236) (3871.3,0.5242) (4062.4,0.5392) (4249.6,0.5443) (4426.7,0.5465) (4612.5,0.5282) (4801.2,0.5382) (5014.4,0.5298) (5241.1,0.5281) (5487.3,0.5281) (5723.0,0.5180) (5973.1,0.5475) (6205.1,0.5502) (6451.2,0.5721) (6691.7,0.5850) (6913.7,0.5869) (7131.9,0.5757) (7341.9,0.5655) (7539.6,0.5825) (7737.3,0.5770) (7970.4,0.5909) (8331.0,0.6054) (8735.3,0.6234) (9161.4,0.6293) (9630.0,0.6454) (10193.1,0.6674) (10789.7,0.6573) (11435.7,0.6638) (12119.3,0.6700) (12822.4,0.6801) (13592.2,0.6894) (14414.0,0.6844) (15299.8,0.6860) (16244.9,0.6956) (17303.7,0.6962) (18451.8,0.7077) (19718.8,0.7230) (21470.3,0.7345) (23401.6,0.7421) (25556.4,0.7596) (27814.0,0.7592) (29867.8,0.7588) (32587.4,0.7665)};
\addplot[ModelUltramarine, fill opacity=0.18, forget plot] fill between[of=U2 and L2];
\addplot[draw=none, forget plot, name path=U3] coordinates {(77.5,0.3026) (83.0,0.2646) (88.8,0.2706) (94.2,0.2584) (100.0,0.2768) (106.1,0.2687) (112.2,0.2782) (118.6,0.2722) (125.0,0.2741) (131.6,0.2765) (138.5,0.2838) (145.8,0.2849) (153.0,0.2484) (160.1,0.2636) (167.0,0.2651) (176.5,0.2817) (185.7,0.2831) (194.8,0.2882) (204.1,0.2900) (214.4,0.2886) (225.5,0.2728) (236.6,0.2705) (248.2,0.2842) (260.0,0.2325) (272.1,0.2444) (284.3,0.2558) (296.7,0.2676) (309.6,0.2877) (325.2,0.2938) (341.5,0.3000) (357.8,0.3027) (374.9,0.3311) (393.8,0.3317) (412.9,0.3206) (432.8,0.3491) (447.6,0.3419) (462.3,0.3633) (477.4,0.3854) (492.2,0.4217) (506.1,0.4217) (519.3,0.4262) (535.2,0.4447) (551.0,0.4363) (566.4,0.4399) (582.7,0.4389) (600.8,0.4646) (617.9,0.4551) (635.4,0.4687) (650.2,0.4891) (667.0,0.5002) (684.0,0.5075) (702.0,0.4925) (722.1,0.5052) (744.5,0.5160) (767.9,0.5121) (792.2,0.5106) (817.9,0.5106) (844.5,0.5011) (872.7,0.4784) (903.3,0.4865) (936.5,0.4922) (968.4,0.4868) (1001.3,0.4887) (1035.2,0.4994) (1070.3,0.5017) (1104.6,0.4753) (1139.6,0.4809) (1176.3,0.4632) (1212.7,0.4563) (1249.5,0.4483) (1288.0,0.4487) (1327.6,0.4821) (1362.9,0.4950) (1397.2,0.4916) (1431.3,0.4809) (1468.4,0.4911) (1507.1,0.4916) (1546.5,0.5012) (1585.3,0.5042) (1622.6,0.5079) (1661.7,0.5433) (1699.3,0.5433) (1737.6,0.5778) (1777.0,0.5929) (1817.0,0.5942) (1861.3,0.5938) (1915.1,0.5997) (1980.0,0.6096) (2046.1,0.6202) (2113.8,0.6250) (2180.8,0.6509) (2248.9,0.6607) (2316.7,0.6478) (2394.5,0.6730) (2477.3,0.6743) (2569.4,0.6813) (2663.6,0.6773) (2769.6,0.6707) (2892.7,0.6788) (3031.4,0.6930) (3176.3,0.6943) (3334.4,0.6952) (3502.5,0.6773) (3682.4,0.6789) (3871.3,0.6789) (4062.4,0.6883) (4249.6,0.6927) (4426.7,0.6889) (4612.5,0.6759) (4801.2,0.6899) (5014.4,0.6719) (5241.1,0.6700) (5487.3,0.6711) (5723.0,0.6511) (5973.1,0.6763) (6205.1,0.6811) (6451.2,0.6967) (6691.7,0.7142) (6913.7,0.7188) (7131.9,0.7072) (7341.9,0.6836) (7539.6,0.7067) (7737.3,0.7076) (7970.4,0.7208) (8331.0,0.7470) (8735.3,0.7558) (9161.4,0.7630) (9630.0,0.7867) (10193.1,0.7939) (10789.7,0.7866) (11435.7,0.7953) (12119.3,0.8042) (12822.4,0.8137) (13592.2,0.8195) (14414.0,0.8127) (15299.8,0.8151) (16244.9,0.8310) (17303.7,0.8320) (18451.8,0.8474) (19718.8,0.8529) (21470.3,0.8591) (23401.6,0.8676) (25556.4,0.8687) (27814.0,0.8663) (29867.8,0.8650) (32587.4,0.8772)};
\addplot[draw=none, forget plot, name path=L3] coordinates {(77.5,0.0857) (83.0,0.0648) (88.8,0.0731) (94.2,0.0733) (100.0,0.0912) (106.1,0.0839) (112.2,0.0961) (118.6,0.0914) (125.0,0.0948) (131.6,0.0989) (138.5,0.1094) (145.8,0.1117) (153.0,0.1050) (160.1,0.1219) (167.0,0.1253) (176.5,0.1441) (185.7,0.1477) (194.8,0.1507) (204.1,0.1540) (214.4,0.1524) (225.5,0.1431) (236.6,0.1377) (248.2,0.1483) (260.0,0.1327) (272.1,0.1345) (284.3,0.1436) (296.7,0.1501) (309.6,0.1600) (325.2,0.1698) (341.5,0.1797) (357.8,0.1830) (374.9,0.2064) (393.8,0.2069) (412.9,0.1902) (432.8,0.2136) (447.6,0.2065) (462.3,0.2293) (477.4,0.2372) (492.2,0.2610) (506.1,0.2610) (519.3,0.2675) (535.2,0.2976) (551.0,0.2857) (566.4,0.2963) (582.7,0.2957) (600.8,0.3088) (617.9,0.2948) (635.4,0.3004) (650.2,0.3196) (667.0,0.3335) (684.0,0.3440) (702.0,0.3276) (722.1,0.3389) (744.5,0.3646) (767.9,0.3620) (792.2,0.3572) (817.9,0.3571) (844.5,0.3525) (872.7,0.3403) (903.3,0.3551) (936.5,0.3623) (968.4,0.3572) (1001.3,0.3628) (1035.2,0.3798) (1070.3,0.3815) (1104.6,0.3640) (1139.6,0.3791) (1176.3,0.3731) (1212.7,0.3709) (1249.5,0.3617) (1288.0,0.3620) (1327.6,0.3810) (1362.9,0.3840) (1397.2,0.3787) (1431.3,0.3646) (1468.4,0.3861) (1507.1,0.3865) (1546.5,0.3903) (1585.3,0.3939) (1622.6,0.3978) (1661.7,0.4088) (1699.3,0.4088) (1737.6,0.4343) (1777.0,0.4442) (1817.0,0.4454) (1861.3,0.4443) (1915.1,0.4526) (1980.0,0.4617) (2046.1,0.4696) (2113.8,0.4799) (2180.8,0.4982) (2248.9,0.5015) (2316.7,0.4901) (2394.5,0.5160) (2477.3,0.5215) (2569.4,0.5303) (2663.6,0.5225) (2769.6,0.5158) (2892.7,0.5261) (3031.4,0.5397) (3176.3,0.5401) (3334.4,0.5425) (3502.5,0.5307) (3682.4,0.5315) (3871.3,0.5315) (4062.4,0.5538) (4249.6,0.5596) (4426.7,0.5554) (4612.5,0.5379) (4801.2,0.5529) (5014.4,0.5422) (5241.1,0.5418) (5487.3,0.5437) (5723.0,0.5288) (5973.1,0.5563) (6205.1,0.5601) (6451.2,0.5816) (6691.7,0.5959) (6913.7,0.6001) (7131.9,0.5875) (7341.9,0.5734) (7539.6,0.5960) (7737.3,0.5975) (7970.4,0.6017) (8331.0,0.6182) (8735.3,0.6273) (9161.4,0.6348) (9630.0,0.6524) (10193.1,0.6723) (10789.7,0.6617) (11435.7,0.6730) (12119.3,0.6761) (12822.4,0.6922) (13592.2,0.7078) (14414.0,0.7019) (15299.8,0.7047) (16244.9,0.7131) (17303.7,0.7138) (18451.8,0.7245) (19718.8,0.7394) (21470.3,0.7551) (23401.6,0.7591) (25556.4,0.7633) (27814.0,0.7620) (29867.8,0.7615) (32587.4,0.7703)};
\addplot[GSMoEOrange, fill opacity=0.18, forget plot] fill between[of=U3 and L3];
\addplot[color=TUMDarkGray] coordinates {(77.5,0.0466) (83.0,0.0508) (88.8,0.0607) (94.2,0.0653) (100.0,0.0745) (106.1,0.0709) (112.2,0.0757) (118.6,0.0776) (125.0,0.0789) (131.6,0.0797) (138.5,0.0921) (145.8,0.0924) (153.0,0.1065) (160.1,0.1124) (167.0,0.1147) (176.5,0.1219) (185.7,0.1222) (194.8,0.1201) (204.1,0.1312) (214.4,0.1341) (225.5,0.1318) (236.6,0.1278) (248.2,0.1374) (260.0,0.1336) (272.1,0.1508) (284.3,0.1634) (296.7,0.1769) (309.6,0.1728) (325.2,0.1768) (341.5,0.1900) (357.8,0.1897) (374.9,0.1978) (393.8,0.2037) (412.9,0.1992) (432.8,0.2190) (447.6,0.2234) (462.3,0.2462) (477.4,0.2667) (492.2,0.2898) (506.1,0.2863) (519.3,0.2865) (535.2,0.3116) (551.0,0.2952) (566.4,0.3052) (582.7,0.2847) (600.8,0.3049) (617.9,0.2921) (635.4,0.3194) (650.2,0.3408) (667.0,0.3511) (684.0,0.3581) (702.0,0.3544) (722.1,0.3601) (744.5,0.3790) (767.9,0.3783) (792.2,0.3726) (817.9,0.3644) (844.5,0.3576) (872.7,0.3368) (903.3,0.3524) (936.5,0.3540) (968.4,0.3437) (1001.3,0.3500) (1035.2,0.3646) (1070.3,0.3839) (1104.6,0.3595) (1139.6,0.3756) (1176.3,0.3702) (1212.7,0.3614) (1249.5,0.3537) (1288.0,0.3607) (1327.6,0.3935) (1362.9,0.4006) (1397.2,0.3937) (1431.3,0.3797) (1468.4,0.3944) (1507.1,0.4060) (1546.5,0.4113) (1585.3,0.4122) (1622.6,0.4061) (1661.7,0.4366) (1699.3,0.4414) (1737.6,0.4742) (1777.0,0.4681) (1817.0,0.4716) (1861.3,0.4697) (1915.1,0.4725) (1980.0,0.4658) (2046.1,0.4839) (2113.8,0.4906) (2180.8,0.5061) (2248.9,0.5128) (2316.7,0.5051) (2394.5,0.5223) (2477.3,0.5290) (2569.4,0.5414) (2663.6,0.5324) (2769.6,0.5288) (2892.7,0.5467) (3031.4,0.5704) (3176.3,0.5707) (3334.4,0.5738) (3502.5,0.5572) (3682.4,0.5578) (3871.3,0.5560) (4062.4,0.5630) (4249.6,0.5700) (4426.7,0.5761) (4612.5,0.5556) (4801.2,0.5677) (5014.4,0.5562) (5241.1,0.5467) (5487.3,0.5501) (5723.0,0.5433) (5973.1,0.5702) (6205.1,0.5739) (6451.2,0.5920) (6691.7,0.6088) (6913.7,0.6111) (7131.9,0.6035) (7341.9,0.5871) (7539.6,0.6007) (7737.3,0.5939) (7970.4,0.6204) (8331.0,0.6418) (8735.3,0.6630) (9161.4,0.6702) (9630.0,0.6924) (10193.1,0.7112) (10789.7,0.7034) (11435.7,0.7079) (12119.3,0.7225) (12822.4,0.7252) (13592.2,0.7301) (14414.0,0.7241) (15299.8,0.7252) (16244.9,0.7393) (17303.7,0.7400) (18451.8,0.7540) (19718.8,0.7629) (21470.3,0.7723) (23401.6,0.7854) (25556.4,0.8053) (27814.0,0.8047) (29867.8,0.8042) (32587.4,0.8144)};
\addlegendentry{DenseNet161Ext}
\addplot[color=Horizon] coordinates {(77.5,0.0000) (83.0,0.0000) (88.8,0.0000) (94.2,0.0000) (100.0,0.0000) (106.1,0.0000) (112.2,0.0000) (118.6,0.0000) (125.0,0.0000) (131.6,0.0000) (138.5,0.0000) (145.8,0.0000) (153.0,0.0000) (160.1,0.0000) (167.0,0.0000) (176.5,0.0000) (185.7,0.0000) (194.8,0.0000) (204.1,0.0019) (214.4,0.0019) (225.5,0.0041) (236.6,0.0041) (248.2,0.0074) (260.0,0.0074) (272.1,0.0333) (284.3,0.0389) (296.7,0.0518) (309.6,0.0518) (325.2,0.0518) (341.5,0.0581) (357.8,0.0669) (374.9,0.0669) (393.8,0.0822) (412.9,0.0822) (432.8,0.1078) (447.6,0.1078) (462.3,0.1260) (477.4,0.1552) (492.2,0.1883) (506.1,0.1883) (519.3,0.1964) (535.2,0.2202) (551.0,0.2169) (566.4,0.2273) (582.7,0.2015) (600.8,0.2309) (617.9,0.2181) (635.4,0.2530) (650.2,0.2804) (667.0,0.2987) (684.0,0.3102) (702.0,0.3184) (722.1,0.3153) (744.5,0.3325) (767.9,0.3346) (792.2,0.3355) (817.9,0.3353) (844.5,0.3329) (872.7,0.3120) (903.3,0.3293) (936.5,0.3335) (968.4,0.3278) (1001.3,0.3389) (1035.2,0.3520) (1070.3,0.3808) (1104.6,0.3524) (1139.6,0.3735) (1176.3,0.3651) (1212.7,0.3563) (1249.5,0.3496) (1288.0,0.3559) (1327.6,0.3878) (1362.9,0.4099) (1397.2,0.4001) (1431.3,0.3849) (1468.4,0.4044) (1507.1,0.4120) (1546.5,0.4169) (1585.3,0.4189) (1622.6,0.4133) (1661.7,0.4420) (1699.3,0.4459) (1737.6,0.4769) (1777.0,0.4726) (1817.0,0.4725) (1861.3,0.4709) (1915.1,0.4712) (1980.0,0.4670) (2046.1,0.4851) (2113.8,0.4917) (2180.8,0.5082) (2248.9,0.5125) (2316.7,0.5044) (2394.5,0.5294) (2477.3,0.5303) (2569.4,0.5447) (2663.6,0.5378) (2769.6,0.5313) (2892.7,0.5474) (3031.4,0.5722) (3176.3,0.5749) (3334.4,0.5758) (3502.5,0.5620) (3682.4,0.5651) (3871.3,0.5639) (4062.4,0.5796) (4249.6,0.5916) (4426.7,0.5984) (4612.5,0.5814) (4801.2,0.5930) (5014.4,0.5804) (5241.1,0.5727) (5487.3,0.5737) (5723.0,0.5677) (5973.1,0.5973) (6205.1,0.5993) (6451.2,0.6187) (6691.7,0.6352) (6913.7,0.6386) (7131.9,0.6265) (7341.9,0.6085) (7539.6,0.6228) (7737.3,0.6154) (7970.4,0.6396) (8331.0,0.6608) (8735.3,0.6784) (9161.4,0.6842) (9630.0,0.7037) (10193.1,0.7187) (10789.7,0.7102) (11435.7,0.7158) (12119.3,0.7286) (12822.4,0.7341) (13592.2,0.7380) (14414.0,0.7323) (15299.8,0.7339) (16244.9,0.7456) (17303.7,0.7465) (18451.8,0.7597) (19718.8,0.7721) (21470.3,0.7825) (23401.6,0.7949) (25556.4,0.8099) (27814.0,0.8092) (29867.8,0.8085) (32587.4,0.8188)};
\addlegendentry{ExplicitMoE}
\addplot[color=ModelUltramarine] coordinates {(77.5,0.0778) (83.0,0.0816) (88.8,0.0862) (94.2,0.1156) (100.0,0.1260) (106.1,0.1175) (112.2,0.1241) (118.6,0.1304) (125.0,0.1324) (131.6,0.1262) (138.5,0.1383) (145.8,0.1411) (153.0,0.1540) (160.1,0.1680) (167.0,0.1600) (176.5,0.1725) (185.7,0.1737) (194.8,0.1689) (204.1,0.1796) (214.4,0.1826) (225.5,0.1825) (236.6,0.1804) (248.2,0.1981) (260.0,0.1699) (272.1,0.1879) (284.3,0.2053) (296.7,0.2180) (309.6,0.2242) (325.2,0.2320) (341.5,0.2430) (357.8,0.2442) (374.9,0.2576) (393.8,0.2634) (412.9,0.2494) (432.8,0.2694) (447.6,0.2712) (462.3,0.2947) (477.4,0.3088) (492.2,0.3362) (506.1,0.3345) (519.3,0.3377) (535.2,0.3611) (551.0,0.3474) (566.4,0.3602) (582.7,0.3508) (600.8,0.3652) (617.9,0.3525) (635.4,0.3759) (650.2,0.3948) (667.0,0.4068) (684.0,0.4160) (702.0,0.4136) (722.1,0.4240) (744.5,0.4436) (767.9,0.4418) (792.2,0.4360) (817.9,0.4306) (844.5,0.4242) (872.7,0.4029) (903.3,0.4117) (936.5,0.4109) (968.4,0.3997) (1001.3,0.4035) (1035.2,0.4154) (1070.3,0.4241) (1104.6,0.4048) (1139.6,0.4176) (1176.3,0.4081) (1212.7,0.3986) (1249.5,0.3887) (1288.0,0.3907) (1327.6,0.4155) (1362.9,0.4228) (1397.2,0.4201) (1431.3,0.4105) (1468.4,0.4246) (1507.1,0.4325) (1546.5,0.4354) (1585.3,0.4411) (1622.6,0.4428) (1661.7,0.4727) (1699.3,0.4800) (1737.6,0.5110) (1777.0,0.5071) (1817.0,0.5072) (1861.3,0.5042) (1915.1,0.5127) (1980.0,0.5133) (2046.1,0.5301) (2113.8,0.5415) (2180.8,0.5608) (2248.9,0.5661) (2316.7,0.5566) (2394.5,0.5747) (2477.3,0.5750) (2569.4,0.5855) (2663.6,0.5765) (2769.6,0.5731) (2892.7,0.5832) (3031.4,0.6016) (3176.3,0.6024) (3334.4,0.6039) (3502.5,0.5876) (3682.4,0.5930) (3871.3,0.5935) (4062.4,0.6033) (4249.6,0.6083) (4426.7,0.6104) (4612.5,0.5939) (4801.2,0.6045) (5014.4,0.5915) (5241.1,0.5866) (5487.3,0.5867) (5723.0,0.5763) (5973.1,0.6051) (6205.1,0.6087) (6451.2,0.6269) (6691.7,0.6421) (6913.7,0.6443) (7131.9,0.6335) (7341.9,0.6172) (7539.6,0.6340) (7737.3,0.6300) (7970.4,0.6493) (8331.0,0.6696) (8735.3,0.6856) (9161.4,0.6920) (9630.0,0.7119) (10193.1,0.7271) (10789.7,0.7187) (11435.7,0.7245) (12119.3,0.7347) (12822.4,0.7424) (13592.2,0.7486) (14414.0,0.7426) (15299.8,0.7442) (16244.9,0.7577) (17303.7,0.7585) (18451.8,0.7725) (19718.8,0.7836) (21470.3,0.7926) (23401.6,0.8026) (25556.4,0.8132) (27814.0,0.8125) (29867.8,0.8117) (32587.4,0.8213)};
\addlegendentry{ImplicitMoE}
\addplot[color=GSMoEOrange] coordinates {(77.5,0.1942) (83.0,0.1647) (88.8,0.1719) (94.2,0.1658) (100.0,0.1840) (106.1,0.1763) (112.2,0.1872) (118.6,0.1818) (125.0,0.1845) (131.6,0.1877) (138.5,0.1966) (145.8,0.1983) (153.0,0.1767) (160.1,0.1927) (167.0,0.1952) (176.5,0.2129) (185.7,0.2154) (194.8,0.2194) (204.1,0.2220) (214.4,0.2205) (225.5,0.2079) (236.6,0.2041) (248.2,0.2163) (260.0,0.1826) (272.1,0.1894) (284.3,0.1997) (296.7,0.2088) (309.6,0.2238) (325.2,0.2318) (341.5,0.2399) (357.8,0.2429) (374.9,0.2688) (393.8,0.2693) (412.9,0.2554) (432.8,0.2813) (447.6,0.2742) (462.3,0.2963) (477.4,0.3113) (492.2,0.3413) (506.1,0.3413) (519.3,0.3468) (535.2,0.3711) (551.0,0.3610) (566.4,0.3681) (582.7,0.3673) (600.8,0.3867) (617.9,0.3750) (635.4,0.3845) (650.2,0.4044) (667.0,0.4168) (684.0,0.4258) (702.0,0.4101) (722.1,0.4220) (744.5,0.4403) (767.9,0.4370) (792.2,0.4339) (817.9,0.4339) (844.5,0.4268) (872.7,0.4094) (903.3,0.4208) (936.5,0.4273) (968.4,0.4220) (1001.3,0.4257) (1035.2,0.4396) (1070.3,0.4416) (1104.6,0.4197) (1139.6,0.4300) (1176.3,0.4181) (1212.7,0.4136) (1249.5,0.4050) (1288.0,0.4053) (1327.6,0.4316) (1362.9,0.4395) (1397.2,0.4351) (1431.3,0.4227) (1468.4,0.4386) (1507.1,0.4391) (1546.5,0.4458) (1585.3,0.4490) (1622.6,0.4529) (1661.7,0.4761) (1699.3,0.4760) (1737.6,0.5061) (1777.0,0.5186) (1817.0,0.5198) (1861.3,0.5191) (1915.1,0.5262) (1980.0,0.5357) (2046.1,0.5449) (2113.8,0.5525) (2180.8,0.5746) (2248.9,0.5811) (2316.7,0.5690) (2394.5,0.5945) (2477.3,0.5979) (2569.4,0.6058) (2663.6,0.5999) (2769.6,0.5933) (2892.7,0.6025) (3031.4,0.6164) (3176.3,0.6172) (3334.4,0.6189) (3502.5,0.6040) (3682.4,0.6052) (3871.3,0.6052) (4062.4,0.6211) (4249.6,0.6262) (4426.7,0.6221) (4612.5,0.6069) (4801.2,0.6214) (5014.4,0.6071) (5241.1,0.6059) (5487.3,0.6074) (5723.0,0.5900) (5973.1,0.6163) (6205.1,0.6206) (6451.2,0.6392) (6691.7,0.6551) (6913.7,0.6594) (7131.9,0.6473) (7341.9,0.6285) (7539.6,0.6514) (7737.3,0.6526) (7970.4,0.6612) (8331.0,0.6826) (8735.3,0.6915) (9161.4,0.6989) (9630.0,0.7196) (10193.1,0.7331) (10789.7,0.7241) (11435.7,0.7342) (12119.3,0.7401) (12822.4,0.7529) (13592.2,0.7636) (14414.0,0.7573) (15299.8,0.7599) (16244.9,0.7720) (17303.7,0.7729) (18451.8,0.7859) (19718.8,0.7961) (21470.3,0.8071) (23401.6,0.8134) (25556.4,0.8160) (27814.0,0.8141) (29867.8,0.8133) (32587.4,0.8238)};
\addlegendentry{GS-MoE}
\end{axis}
\end{tikzpicture}

%% file: figures/scatter.tex
\begin{tikzpicture}

\pgfplotsset{
    base scatter/.style={
        width=0.31\textwidth,
        height=0.31\textwidth,
        xlabel style={font=\tiny, yshift=1mm}, 
        ylabel style={font=\tiny, yshift=-1mm},
        xticklabel style={font=\tiny},
        yticklabel style={font=\tiny},
        xmin=0.0, xmax=1.0,
        ymin=0.0, ymax=1.0,
        xtick={0.2,0.4,0.6,0.8,1.0},
        ytick={0.2,0.4,0.6,0.8,1.0},
        grid=both,
        minor grid style={gray!10},
        major grid style={gray!20},
        title style={font=\footnotesize, at={(0.5,0.95)}, anchor=south},
    }
}

\begin{axis}[
    base scatter,
    name=plotA,
    at={(0,0)},
    anchor=south west,
    xlabel={DenseNet161Ext F1-Score},
    ylabel={Explicit MoE F1-Score},
    legend style={
        draw=none,
        font=\tiny, 
        at={(1.85, 1.15)}, 
        anchor=south,
        fill=none,
        row sep=-2pt,
        legend columns=7, 
    },
    legend cell align=left,
    title={\tiny\textbf{(a)}},
]
\addplot[gray, dashed, domain=0:1, samples=2, forget plot] {x};

\addplot[Cerulean, fill=Cerulean, mark=diamond*, only marks, mark size=1.8pt, opacity=0.8]
coordinates {(0.1935,0.1765) (0.7012,0.7328) (0.7344,0.7424) (0.5847,0.6088) (0.6107,0.6494) (0.6883,0.7240) (0.0000,0.0000) (0.4457,0.3854) (0.2857,0.0000) (0.1176,0.0000) (0.4527,0.4488) (0.2500,0.0000) (0.2025,0.0000) (0.0000,0.0000) (0.7057,0.7616) (0.3809,0.4190) (0.1803,0.0000) (0.3713,0.4463) (0.7044,0.7312) (0.2709,0.2360) (0.3411,0.4191) (0.6181,0.6533) (0.7037,0.7211) (0.0000,0.0000) (0.1765,0.0000)};
\addlegendentry{MR-AF}

\addplot[Horizon, fill=Horizon, mark=diamond*, only marks, mark size=2pt, opacity=0.8]
coordinates {(0.5091,0.5169) (0.2564,0.0000) (0.4101,0.4082) (0.0000,0.0000) (0.5291,0.2442) (0.2914,0.0000) (0.5015,0.4700) (0.9860,0.9892) (0.0000,0.0000) (0.8702,0.8567)};
\addlegendentry{MR-Brain}

\addplot[ModelUltramarine, fill=ModelUltramarine, mark=diamond*, only marks, mark size=1.8pt, opacity=0.7]
coordinates {(0.3012,0.1266) (0.0000,0.0000) (0.2256,0.2062) (0.0000,0.0000) (0.2000,0.0000) (0.3636,0.0000) (0.3636,0.0000) (0.0000,0.0000) (0.1159,0.0000) (0.0000,0.0000) (0.3182,0.0000) (0.2121,0.0444) (0.0175,0.0000) (0.4388,0.4930) (0.4836,0.4258) (0.3883,0.3636) (0.9119,0.9158) (0.3167,0.2569) (0.5060,0.5328) (0.3038,0.4072) (0.0000,0.0000) (0.4719,0.5333) (0.0000,0.0000) (0.0513,0.0000) (0.0000,0.0000) (0.7123,0.7326)};
\addlegendentry{MR-Abd}

\addplot[Endeavour, fill=white, mark=diamond*, only marks, mark size=1.8pt, opacity=0.6]
coordinates {(0.2456,0.2087) (0.0000,0.0000) (0.5000,0.5834) (0.2143,0.0000) (0.1091,0.0385) (0.7769,0.7709) (0.4636,0.4890) (0.5292,0.5660) (0.1928,0.0000) (0.2836,0.1353) (0.4408,0.3226) (0.2324,0.1479) (0.4810,0.5232) (0.0260,0.0000) (0.7352,0.7356) (0.6003,0.6117) (0.7335,0.7673) (0.1622,0.0000) (0.3750,0.3467) (0.1176,0.0000) (0.3043,0.2660) (0.6118,0.6667) (0.7234,0.7365) (0.2667,0.1132) (0.6077,0.6953) (0.4647,0.5372) (0.4219,0.5122) (0.1818,0.0000) (0.3409,0.0656) (0.4914,0.5753) (0.6833,0.6932) (0.0714,0.0000) (0.3532,0.4378) (0.3714,0.2900) (0.0000,0.0000) (0.1406,0.1045) (0.0000,0.0000) (0.7716,0.7828) (0.5036,0.5467) (0.8160,0.8425) (0.5395,0.4757) (0.3662,0.3432) (0.6461,0.6450) (0.5465,0.5481) (0.0000,0.0000) (0.4744,0.5052) (0.2800,0.3048) (0.1159,0.0177) (0.7274,0.7423) (0.3220,0.3939) (0.1333,0.0000) (0.5870,0.5599) (0.8163,0.8357) (0.5322,0.5111) (0.7177,0.8023)};
\addlegendentry{Other MR}

\addplot[GSMoRed, fill=GSMoRed, mark=*, only marks, mark size=1.5pt, opacity=0.8]
coordinates {(0.9670,0.9728) (0.6142,0.5854) (0.8526,0.8646) (0.7047,0.6854) (0.9702,0.9772) (0.6667,0.6994) (0.9673,0.9772) (0.9420,0.9573) (0.8826,0.8870) (0.9696,0.9736) (0.7000,0.7000) (0.8873,0.9196) (0.9751,0.9909) (0.9159,0.9295) (0.9880,0.9959)};
\addlegendentry{US}

\addplot[GSMoEYellow, fill=GSMoEYellow, mark=*, only marks, mark size=1.3pt, opacity=0.8]
coordinates {(0.8453,0.8532) (0.2135,0.2222) (0.5493,0.5501) (0.2170,0.3703) (0.4661,0.5299) (0.7575,0.7673)};
\addlegendentry{CT-Lung}

\addplot[GSMoEOrange, fill=GSMoEOrange, mark=*, only marks, mark size=1.3pt, opacity=0.7]
coordinates {(0.0000,0.0000) (0.3420,0.3578) (0.4527,0.5537) (0.2703,0.2825) (0.0000,0.0000) (0.4516,0.4207) (0.5834,0.5435) (0.4232,0.4560) (0.0000,0.0000) (0.0000,0.0000) (0.1026,0.1630) (0.0000,0.0000) (0.0000,0.0000) (0.6291,0.6353) (0.0000,0.0000) (0.6867,0.7138) (0.8199,0.7198) (0.8938,0.8641) (0.3272,0.3600) (0.2507,0.2488) (0.3580,0.3736) (0.7745,0.7016) (0.5664,0.4877) (0.0000,0.0000) (0.1875,0.0000) (0.0833,0.0000) (0.3410,0.3614) (0.5575,0.5721)};
\addlegendentry{CT-Abd}

\end{axis}

\begin{axis}[
    base scatter,
    name=plotB,
    at={(plotA.south east)},
    anchor=south west,
    xshift=8mm, 
    xlabel={DenseNet161Ext F1-Score},
    ylabel={GS-MoE F1-Score},
    title={\tiny\textbf{(b)}},
]
\addplot[gray, dashed, domain=0:1, samples=2, forget plot] {x};

\addplot[Cerulean, fill=Cerulean, mark=diamond*, only marks, mark size=1.8pt, opacity=0.8]
coordinates {(0.1935,0.2697) (0.7012,0.8532) (0.7344,0.8646) (0.5847,0.6814) (0.6107,0.8185) (0.6883,0.7801) (0.0000,0.8571) (0.4457,0.7140) (0.2857,0.6015) (0.1176,0.2182) (0.4527,0.6647) (0.2500,0.8070) (0.2025,0.4731) (0.0000,0.4615) (0.7057,0.6907) (0.3809,0.5799) (0.1803,0.4512) (0.3713,0.7034) (0.7044,0.7652) (0.2709,0.4837) (0.3411,0.5897) (0.6181,0.7634) (0.7037,0.7373) (0.0000,0.6667) (0.1765,0.5517)};

\addplot[Horizon, fill=Horizon, mark=diamond*, only marks, mark size=2pt, opacity=0.8]
coordinates {(0.5091,0.4771) (0.2564,0.2500) (0.4101,0.4797) (0.0000,0.4167) (0.5291,0.5565) (0.2914,0.2174) (0.5015,0.5348) (0.9860,0.9890) (0.0000,0.1818) (0.8702,0.8581)};

\addplot[ModelUltramarine, fill=ModelUltramarine, mark=diamond*, only marks, mark size=1.8pt, opacity=0.7]
coordinates {(0.3012,0.3190) (0.0000,0.0000) (0.2256,0.3478) (0.0000,0.2609) (0.2000,0.3600) (0.3636,0.3111) (0.3636,0.2632) (0.0000,0.0000) (0.1159,0.0769) (0.0000,0.0000) (0.3182,0.2381) (0.2121,0.2791) (0.0175,0.1547) (0.4388,0.4302) (0.4836,0.4971) (0.3883,0.5468) (0.9119,0.9024) (0.3167,0.2958) (0.5060,0.4354) (0.3038,0.4000) (0.0000,0.5714) (0.4719,0.4270) (0.0000,0.2143) (0.0513,0.0370) (0.0000,0.0789) (0.7123,0.7339)};

\addplot[Endeavour, fill=white, mark=diamond*, only marks, mark size=1.8pt, opacity=0.6]
coordinates {(0.2456,0.2299) (0.0000,0.1176) (0.5000,0.5293) (0.2143,0.2500) (0.1091,0.2128) (0.7769,0.7754) (0.4636,0.4929) (0.5292,0.5711) (0.1928,0.1301) (0.2836,0.3045) (0.4408,0.4701) (0.2324,0.3271) (0.4810,0.5122) (0.0260,0.0220) (0.7352,0.7157) (0.6003,0.6015) (0.7335,0.7635) (0.1622,0.2025) (0.3750,0.3839) (0.1176,0.2424) (0.3043,0.3450) (0.6118,0.6241) (0.7234,0.7317) (0.2667,0.3200) (0.6077,0.6506) (0.4647,0.5425) (0.4219,0.5000) (0.1818,0.0000) (0.3409,0.5055) (0.4914,0.5245) (0.6833,0.6807) (0.0714,0.0000) (0.3532,0.4057) (0.3714,0.3130) (0.0000,0.0000) (0.1406,0.2839) (0.0000,0.0000) (0.7716,0.7692) (0.5036,0.5882) (0.8160,0.8473) (0.5395,0.5079) (0.3662,0.3804) (0.6461,0.6328) (0.5465,0.5596) (0.0000,0.0000) (0.4744,0.5405) (0.2800,0.2778) (0.1159,0.2079) (0.7274,0.7261) (0.3220,0.3964) (0.1333,0.1212) (0.5870,0.5459) (0.8163,0.8247) (0.5322,0.5139) (0.7177,0.7143)};

\addplot[GSMoRed, fill=GSMoRed, mark=*, only marks, mark size=1.5pt, opacity=0.8]
coordinates {(0.9670,0.9664) (0.6142,0.6324) (0.8526,0.8674) (0.7047,0.7065) (0.9702,0.9777) (0.6667,0.7020) (0.9673,0.9763) (0.9420,0.9574) (0.8826,0.8858) (0.9696,0.9745) (0.7000,0.7368) (0.8873,0.9120) (0.9751,0.9976) (0.9159,0.9239) (0.9880,0.9989)};

\addplot[GSMoEYellow, fill=GSMoEYellow, mark=*, only marks, mark size=1.3pt, opacity=0.8]
coordinates {(0.8453,0.8566) (0.2135,0.2564) (0.5493,0.5613) (0.2170,0.5218) (0.4661,0.5293) (0.7575,0.7554)};

\addplot[GSMoEOrange, fill=GSMoEOrange, mark=*, only marks, mark size=1.3pt, opacity=0.7]
coordinates {(0.0000,0.1176) (0.3420,0.3328) (0.4527,0.5305) (0.2703,0.3774) (0.0000,0.2675) (0.4516,0.3868) (0.5834,0.5191) (0.4232,0.4242) (0.0000,0.0000) (0.0000,0.0000) (0.1026,0.2524) (0.0000,0.0583) (0.0000,0.0000) (0.6291,0.6523) (0.0000,0.0976) (0.6867,0.7004) (0.8199,0.6357) (0.8938,0.8929) (0.3272,0.4730) (0.2507,0.2879) (0.3580,0.3846) (0.7745,0.6967) (0.5664,0.5412) (0.0000,0.0000) (0.1875,0.0674) (0.0833,0.2653) (0.3410,0.3494) (0.5575,0.5826)};

\end{axis}

\begin{axis}[
    base scatter,
    name=plotC,
    at={(plotB.south east)},
    anchor=south west,
    xshift=8mm, 
    xlabel={ExplicitMoE F1-Score},
    ylabel={GS-MoE F1-Score},
    title={\tiny\textbf{(c)}},
]
\addplot[gray, dashed, domain=0:1, samples=2, forget plot] {x};

\addplot[Cerulean, fill=Cerulean, mark=diamond*, only marks, mark size=1.8pt, opacity=0.8]
coordinates {(0.1765,0.2697) (0.7328,0.8532) (0.7424,0.8646) (0.6088,0.6814) (0.6494,0.8185) (0.7240,0.7801) (0.0000,0.8571) (0.3854,0.7140) (0.0000,0.6015) (0.0000,0.2182) (0.4488,0.6647) (0.0000,0.8070) (0.0000,0.4731) (0.0000,0.4615) (0.7616,0.6907) (0.4190,0.5799) (0.0000,0.4512) (0.4463,0.7034) (0.7312,0.7652) (0.2360,0.4837) (0.4191,0.5897) (0.6533,0.7634) (0.7211,0.7373) (0.0000,0.6667) (0.0000,0.5517)};

\addplot[Horizon, fill=Horizon, mark=diamond*, only marks, mark size=2pt, opacity=0.8]
coordinates {(0.5169,0.4771) (0.0000,0.2500) (0.4082,0.4797) (0.0000,0.4167) (0.2442,0.5565) (0.0000,0.2174) (0.4700,0.5348) (0.9892,0.9890) (0.0000,0.1818) (0.8567,0.8581)};

\addplot[ModelUltramarine, fill=ModelUltramarine, mark=diamond*, only marks, mark size=1.8pt, opacity=0.7]
coordinates {(0.1266,0.3190) (0.0000,0.0000) (0.2062,0.3478) (0.0000,0.2609) (0.0000,0.3600) (0.0000,0.3111) (0.0000,0.2632) (0.0000,0.0000) (0.0000,0.0769) (0.0000,0.0000) (0.0000,0.2381) (0.0444,0.2791) (0.0000,0.1547) (0.4930,0.4302) (0.4258,0.4971) (0.3636,0.5468) (0.9158,0.9024) (0.2569,0.2958) (0.5328,0.4354) (0.4072,0.4000) (0.0000,0.5714) (0.5333,0.4270) (0.0000,0.2143) (0.0000,0.0370) (0.0000,0.0789) (0.7326,0.7339)};

\addplot[Endeavour, fill=white, mark=diamond*, only marks, mark size=1.8pt, opacity=0.6]
coordinates {(0.2087,0.2299) (0.0000,0.1176) (0.5834,0.5293) (0.0000,0.2500) (0.0385,0.2128) (0.7709,0.7754) (0.4890,0.4929) (0.5660,0.5711) (0.0000,0.1301) (0.1353,0.3045) (0.3226,0.4701) (0.1479,0.3271) (0.5232,0.5122) (0.0000,0.0220) (0.7352,0.7157) (0.6117,0.6015) (0.7673,0.7635) (0.0000,0.2025) (0.3467,0.3839) (0.0000,0.2424) (0.2660,0.3450) (0.6667,0.6241) (0.7365,0.7317) (0.1132,0.3200) (0.6953,0.6506) (0.5372,0.5425) (0.5122,0.5000) (0.0000,0.0000) (0.0656,0.5055) (0.5753,0.5245) (0.6932,0.6807) (0.0000,0.0000) (0.4378,0.4057) (0.2900,0.3130) (0.0000,0.0000) (0.1045,0.2839) (0.0000,0.0000) (0.7828,0.7692) (0.5467,0.5882) (0.8425,0.8473) (0.4757,0.5079) (0.3432,0.3804) (0.6450,0.6328) (0.5481,0.5596) (0.0000,0.0000) (0.5052,0.5405) (0.3048,0.2778) (0.0177,0.2079) (0.7423,0.7261) (0.3939,0.3964) (0.0000,0.1212) (0.5599,0.5459) (0.8357,0.8247) (0.5111,0.5139) (0.8023,0.7143)};

\addplot[GSMoRed, fill=GSMoRed, mark=*, only marks, mark size=1.5pt, opacity=0.8]
coordinates {(0.9728,0.9664) (0.5854,0.6324) (0.8646,0.8674) (0.6854,0.7065) (0.9772,0.9777) (0.6994,0.7020) (0.9772,0.9763) (0.9573,0.9574) (0.8870,0.8858) (0.9736,0.9745) (0.7000,0.7368) (0.9196,0.9120) (0.9909,0.9976) (0.9295,0.9239) (0.9959,0.9989)};

\addplot[GSMoEYellow, fill=GSMoEYellow, mark=*, only marks, mark size=1.3pt, opacity=0.8]
coordinates {(0.8532,0.8566) (0.2222,0.2564) (0.5501,0.5613) (0.3703,0.5218) (0.5299,0.5293) (0.7673,0.7554)};

\addplot[GSMoEOrange, fill=GSMoEOrange, mark=*, only marks, mark size=1.3pt, opacity=0.7]
coordinates {(0.0000,0.1176) (0.3578,0.3328) (0.5537,0.5305) (0.2825,0.3774) (0.0000,0.2675) (0.4207,0.3868) (0.5435,0.5191) (0.4560,0.4242) (0.0000,0.0000) (0.0000,0.0000) (0.1630,0.2524) (0.0000,0.0583) (0.0000,0.0000) (0.6353,0.6523) (0.0000,0.0976) (0.7138,0.7004) (0.7198,0.6357) (0.8641,0.8929) (0.3600,0.4730) (0.2488,0.2879) (0.3736,0.3846) (0.7016,0.6967) (0.4877,0.5412) (0.0000,0.0000) (0.0000,0.0674) (0.0000,0.2653) (0.3614,0.3494) (0.5721,0.5826)};
\end{axis}

\end{tikzpicture}